\documentclass[journal]{IEEEtran}
\ifCLASSINFOpdf
\else
\fi

\usepackage{hyperref}
\usepackage{float}
\usepackage{ulem}
\usepackage{booktabs}
\usepackage{amssymb}
\usepackage{graphicx}
\usepackage{indentfirst}
\usepackage{subfigure}
\usepackage{algorithm}
\usepackage{algorithmic}
\usepackage{amsmath}
\usepackage{multirow}
\usepackage{threeparttable}  
\usepackage{url}
\usepackage{bibentry}
\usepackage{epsfig}
\usepackage{color}
\usepackage{epstopdf}
\usepackage{rotating}
\usepackage{caption}
\usepackage{multirow}
\usepackage{stfloats} 
\usepackage{color}
\usepackage{bm}
\usepackage{graphicx}
\usepackage{ntheorem}
\usepackage{fancybox}
\usepackage{tikz}
\usetikzlibrary{calc,3d}
\begin{document}
%
\title{Tensor Completion via Convolutional Sparse Coding Regularization}

\author{Zhebin~Wu,
	Tianchi~Liao,
	Chuan~Chen*,
	Cong~Liu,
	Zibin~Zheng,
	and~Xiongjun~Zhang~
	\thanks{Zhebin Wu, Chuan Chen, Cong Liu, and Zibin Zheng were with the School of Data and Computer Science, Sun Yat-sen University, Guangzhou 510006, China. In the meanwhile, they study in National Engineering Research Center of Digital Life, Sun Yat-sen University, Guangzhou, China (e-mail: wuzhb6@mail2.sysu.edu.cn, chenchuan@mail.sysu.edu.cn, liucong3@mail.sysu.edu.cn, zhzibin@mail.sysu.edu.cn).}
	\thanks{Tianchi Liao was with the school of software engineering, Sun Yat-sen University, Zhuhai 519082, China (e-mail: tiaerc@163.com). }
	\thanks{Xiongjun Zheng was with the School of Mathematics and Statistics and Hubei Key Laboratory of Mathematical Sciences, Central China Normal University, Wuhan 430079, China (e-mail: xjzhang@mail.ccnu.edu.cn).}
	\thanks{Chuan Chen is the corresponding author (e-mail: chenchuan@mail.sysu.edu.cn).}
}

\maketitle

\begin{abstract}
Tensor data often suffer from missing value problem due to the complex high-dimensional structure while acquiring them. To complete the missing information, lots of Low-Rank Tensor Completion (LRTC) methods have been proposed, most of which depend on the low-rank property of tensor data. In this way, the low-rank component of the original data could be recovered roughly. However, the shortcoming is that the detail information can not be fully restored, no matter the Sum of the Nuclear Norm (SNN) nor the Tensor Nuclear Norm (TNN) based methods. On the contrary, in the field of signal processing, Convolutional Sparse Coding (CSC) can provide a good representation of the high-frequency component of the image, which is generally associated with the detail component of the data. Nevertheless, CSC can not handle the low-frequency component well. 
To this end, we propose two novel methods, LRTC-CSC-I and LRTC-CSC-II, which adopt CSC as a supplementary regularization for LRTC to capture the high-frequency components.  Therefore, the LRTC-CSC methods can not only solve the missing value problem but also recover the details. Moreover, the regularizer CSC can be trained with small samples due to the sparsity characteristic.
Extensive experiments show the effectiveness of LRTC-CSC methods, and quantitative evaluation indicates that the performance of our models are superior to state-of-the-art methods.


\end{abstract}


%
\IEEEpeerreviewmaketitle

\section{Introduction}
A tensor is often known as an extension of a 1D vector or 2D matrix, which can offer a high-dimensional storage structure for various data nowadays, such as color image, hyperspectral image, video, etc. Thus, it's widely leveraged in the field of computer vision, data mining \cite{chen2018heterogeneous}, machine learning \cite{chen2019tensor}, and large-scale data analysis.
However, it's common that the obtained tensor data is nevertheless undersampled when processing it, which derives many Low-Rank Tensor Completion (LRTC) \cite{liu2012tensor}\cite{gandy2011tensor}\cite{xu2013parallel} algorithms to address the missing value problem. The general LRTC algorithm is formulated as:
\begin{equation}
	\min_{\mathcal{X}} \quad \mathrm{rank}\left(\mathcal{X}\right)\quad \mathrm{s.t.}\quad \mathcal{X}_{\Omega}=\mathcal{T}_{\Omega},
	\label{eqn:lrtc0}
\end{equation}
where $\mathcal{X}$ is the underlying tensor, $\mathcal{T}$ is the original data, $\Omega$ is the index set which implies the location corresponding to the observed entries. However, it's intractable to solve problem (\ref{eqn:lrtc0}) which is NP-hard \cite{kolda2009tensor}, due to the combinational nature of the function rank($\cdot$). 

The definition of tensor rank is different from the known matrix's, because it is not unique. It depends on the tensor decomposition method used, e.g., CANDECOMP/PARAFAC (CP) \cite{carroll1970analysis}, Tucker \cite{tucker1966some}, Tensor-Train (TT) \cite{oseledets2011tensor}, Tensor Ring \cite{zhao2016tensor} etc. The definition of tensor rank, CP rank specifically, was first proposed by Kruskal \textit{et al.} \cite{kruskal1977three}. Immediately following, Tucker rank was adopted to approach the LRTC and gained unexpected achievements \cite{liu2012tensor}\cite{xu2013parallel}. However, the drawback of Tucker rank is that its components will be the ranks of highly unbalanced flattened matrices if the original tensor is high dimensional (N$\textgreater$3). To address above problem, A. Bengua \cite{bengua2017efficient} proposed a novel LRTC solver based on TT rank. Besides, they employ the Ket Augmentation (KA) to extend low dimensional tensors to higher ones and proves that TT rank is more capable of global information capturing when the dimension of a tensor is larger than three. Another novel definition of tensor rank named tubal rank \cite{zhang2014novel}\cite{lu2019tensor} is based on the recently proposed t-SVD decomposition \cite{kilmer2013third}, which is able to characterize the inherent low-rank structure of a tensor. Actually, no matter what rank is exploited to recover corrupted tensors, the core of these LRTC models is to find the relationship between missing entries and retained ones. 

Despite the tensor rank estimation issue is intractable, nuclear norm \cite{recht2010guaranteed} has been proved to be the most effective convex surrogate for the function rank($\cdot$) of the matrix. Liu \textit{et al.} \cite{liu2012tensor} proposed the sum of the nuclear norm (SNN) as a relaxation of the  tensor rank.
In addition, the tensor nuclear norm (TNN) \cite{lu2018exact} is proposed from t-product to keep consistent with the matrix cases on concepts (see details in Section II). No matter which nuclear norm is, the strong prior of the underlying tensor data is its low-rank property, and then the low-rank component can be recovered based on that. 

Inevitably, the details of observation are overlooked by LRTC models. Thus, lots of regularizations are proposed to be employed as another prior to improve the details of recovered data, e.g., Total Variation (TV) \cite{rudin1992nonlinear}\cite{ji2016tensor}, framelets \cite{jiang2018matrix}, wavelets \cite{chen2006oblique} etc. Among these priors, TV is the most popular one in LRTC problems.
Li \textit{et al.} \cite{li2017low} combined TV and low-rankness to recover visual tensors and proposed a novel LRTC-TV model, where TV accounts for local piecewise priors. The starting point of LRTC-TV lies on the fact that the objects or edges in the spatial dimension will hold a smooth and piecewise structure. As a complementary prior to low rankness, TV can explore the local features properly. While Jiang \textit{et al.} \cite{jiang2018anisotropic} combined TNN and an anisotropic TV to propose a TNN-3DTV model to exploit the correlations among the spatial and channel domains. The only fly in the ointment is that TV assumes the underlying tensor is piecewise smooth, resulting in undesired patch effects on the recovered tensors \cite{liu2015image}. What's more, the regularizations above are all only based on the information within that image, which is fatal when the remaining entries are few. Zhang \textit{et al.} \cite{zhang2017learning} utilized Convolutional Neural Networks (CNN) to train a set of effective denoisers to solving the image denoising problem, which shows the importance of feature collections. Thus, it's necessary to introduce other prior information outer the image for a better result. 

Another prominent paradigm in signal processing is sparse coding. By 
seeking a sparse representation under an overcomplete dictionary, the underlying data could be restored exceptionally. The fundamental of sparse coding is the dictionary learning, a process to obtain a dictionary $D$ that can best represent a set of train data. As early as 2006, Elad \textit{et al.} \cite{elad2006image} leveraged dictionary learning-based method to tackle the image denoising problem. Later in 2010, Yang \textit{et al.} \cite{yang2010image} decided to sparsely code for each patch of the low-resolution input and then used the coefficients of learned representation to generate the high-resolution output. In this way, the image super-resolution could be represented by sparse entries. 

For a fixed dictionary $D\in \mathbb{R}^{M\times N}$, given a signal $X\in \mathbb{R}^{N}$, the task of seeking its sparsest representation $\Gamma \in \mathbb{R}^{M}$ is called sparse coding. Mathematically, the sparse coding problem can be formulated as:
\begin{equation}
	\min_{\Gamma}\quad \left\Vert \Gamma \right\Vert_0 \quad \mathrm{s.t.}\quad D\Gamma=X,
	\label{pb:sr}
\end{equation}
where $\left\Vert \Gamma \right\Vert_0$ denotes the number of non-zeros in $\Gamma$. There exists a convex relaxation of problem (\ref{pb:sr}) in the form of Basis-Pursuit (BP) problem \cite{chen2001atomic}, formulated as:
\begin{equation}
	\min_{\Gamma}\quad \left\Vert \Gamma \right\Vert_1 \quad \mathrm{s.t.}\quad D\Gamma=X.
	\label{pb:bp}
\end{equation}
Correspondingly, there are several practical algorithms to solve problem (\ref{pb:bp}), e.g., Orthogonal Matching Pursuit (OMP) \cite{chen1989orthogonal} and the BP \cite{daubechies2004iterative}\cite{chen2001atomic}. However, most traditional dictionary learning methods are patch-based, which might ignore the consistency among the entries lied in different patches but have some common features in the image. Besides, the learned data will contain shifted versions of the same features, resulting in an over-redundant dictionary.

To tackle above problems, an alternative model, Convolutional Sparse Coding (CSC), has gained great attention in machine learning for image and video processing \cite{papyan2017convolutional,heide2015fast} recently due to its shift-invariance property. In 2010, Zeiler \textit{et al.} \cite{zeiler2010deconvolutional} first proposed the concept of CSC. To improve the computional effectiveness, Bristow \textit{et al.} \cite{bristow2013fast} and Heide \textit{et al.} \cite{heide2015fast} proposed novel and fast solutions under the Alternating Direction Method of Multipliers (ADMM) framework in succession.
Furthermore, Wohlberg \cite{wohlberg2018convolutional} demonstrated that suitable penalties on the gradients of the coefficient maps will improve the impulse noise denoising performance. Papyan \textit{et al.} \cite{papyan2017convolutional2} introduced the relationship between CSC and CNN and claimed that CNN can be analyzed theoretically via multi-layer CSC.

In terms of applications, Papyan \textit{et al.}  \cite{papyan2017convolutional} also proposed a slice-based dictionary learning algorithm, which was utilized in the CSC model to inpaint image as well as separate texture and cartoon. Recently, Zhang \textit{et al.} \cite{zhang2017convolutional} integrated the concept of low-rankness into CSC and addressed the rain streak removal. The work of Bao \textit{et al.} \cite{bao2019convolutional} demonstrated that the CSC is able to process the high-frequency component of an image. In order to improve the capacity of the dictionary for learning multi-dimensional data, Bibi \textit{et al.} \cite{bibi2017high} leveraged high order algebra to allow traditional CSC models encoding tensors. In addition, Xu \textit{et al.} \cite{xu2020factorized} decomposed the tensor dictionary and reconstructed it under the orthogonality-constrained convolutional factorization scheme to reduce computation costs. 
All in all, it can be argued that the success of CSC may attribute to the idea, ``think globally and work locally''.

Actually, the regularizations like TV, framelets, and wavelets only play the main role in a single image, which can be treated as forcing a hard constraint of physics on underlying data. The CNN-based regularization is often trained with thousands of data, which can be explained by feature collections in statistics. However, in real world, it's not always available to acquire such a large amount of samples in a limited time. Thus, we take CSC into consideration for its conveniency, which can be trained  only with few-shot samples. The CSC is to extract the optimal representation dictionary in a limited training set. And this is different from the Deep learning (CNN etc.) based approaches, where the statistical feature should be learned from lots of samples. 

Inspired by the capacity of CSC on extracting features locally with an overcomplete dictionary, two novel LRTC-CSC models are proposed to address the LRTC problem. By regarding CSC as a plug-and-play submodel in the optimization framework of the whole model, we effectively augment the high-frequency component of the underlying tensor.
In this way, the global structure is handled by LRTC prior, and then CSC is leveraged to make up for another detail prior. Eventually, both global and local features can be well recovered. The main contributions of this paper are:
\begin{enumerate}
	\item To tackle the missing value problem, we proposed two CSC regularized LRTC models, SNN-based and TNN-based, respectively. In both models, firstly, an overcomplete dictionary is pre-trained only with very small amount of data. And then the dictionary is used in CSC to restore the details of underlying tensor. As a result, both low-rank component and details are well recovered.
	
	\item We came up with effective algorithms to solve the LRTC-CSC-I and LRTC-CSC-II model, which are based on the inexact ADMM method \cite{sutor2015alternating} and plug-and-play framework \cite{venkatakrishnan2013plug}. With the variable splitting techniques, the entire problem can be split into three subproblems and solved separately.
	
	\item We tested our model on different kinds of datasets, including color images, MRI data and videos.
	Extensive experiments have verified the effectiveness of our model and claimed that the CSC regularization is suitable for different LRTC models. Furthermore, the performance of LRTC-CSC-II is superior to state-of-the-art models in the experiments.
\end{enumerate}

\section{Notations and Preliminaries}
\subsection{Notations}

Throughout this paper, the $({i_1i_2\cdots i_n})$-th element of a tensor $\mathcal{X} \in \mathbb{R}^{I_1\times I_2\times \cdots \times I_n}$ is denoted by $x_{i_1i_2\cdots i_n}$.Vectors are denoted by boldface lowercase letters, e.g. $\bm{\alpha}$ and scalars by lowercase letters, e.g. $\alpha$. We denote an $n$-mode tensor by calligraphic letters, e.g., $\mathcal{X} \in \mathbb{R}^{I_1\times I_2\times \cdots \times I_n}$, where $I_k,k=1,2,\cdots,n$ is the dimension of $k$-th mode and $\mathbb{R}$ is the field of real number. Let the upper case latters, e.g., $X$, denote the matrices. Especially, a mode-$k$ matricization (also known as mode-k unfolding or flattening) of a tensor $\mathcal{X}$ is reshaping the tensor into a matrix $X_{(k)} \in \mathbb{R}^{I_k \times(I_1\cdots I_{k-1}I_{k+1}\cdots I_N)}$, which is defined as Tucker rank. For a 3-way tensor $\mathcal{X} \in \mathbb{R}^{n_1\times n_2\times n_3}$, we denote its $(i,j)$-th mode-1, mode-2, and mode-3 fibers as $\mathcal{X}(:,i,j)$, $\mathcal{X}(i,:,j)$, and $\mathcal{X}(i,j,:)$. We use the Matlab notation $\mathcal{X}(i,:,:)$, $\mathcal{X}(:,i,:)$, and $\mathcal{X}(:,:,i)$ to denote the $i$-th horizontal, lateral and frontal slices,  respectively. More often, $X^{(i)}$ and tube are used to represent $\mathcal{X}(:,:,i)$ and mode-3 fiber, respectively.

In addition, there are some mathematical operations of matrices and tensors used in this paper. The inner product of $X\in \mathbb{R}^{n\times m}$ and $Y \in \mathbb{R}^{n\times m}$ is defined as $\langle X,Y\rangle = tr(X^HY)$, where $X^H$ is the conjugate transpose of the matrix $X$ and $tr(\cdot)$ is matrix trace. The $\mathit{l}_1$-norm is defined as $\lVert\mathcal{X}\rVert_1 = \sum_{i_1i_2\cdots i_n} |x_{i_1i_2\cdots i_n}|$, the Frobenius norm as $\lVert\mathcal{X}\rVert_F = \sqrt{\sum_{i_1i_2\cdots i_n} (x_{i_1i_2\cdots i_n})^2}$ and the nuclear norm of matrix as $\lVert X\rVert_\ast = \sum_{i}\sigma_i$, where $\sigma_i$ is the $i$-th largest singular value of matrix $X$. For a vector $\bm{\alpha}$, the $\mathit{l}_2$-norm is $\lVert \bm{\alpha}\rVert_2 =\sqrt{\sum_{i} {\alpha}_i^2}$. For a tensor $\mathcal{X} \in \mathbb{R}^{n_1\times n_2\times n_3}$, we use $\hat{\mathcal{X}}$ to denote the result of discrete Fourier transformation of $\mathcal{X}$ along the 3-rd dimension, i.e., $\hat{\mathcal{X}}=\mathrm{fft}(\mathcal{X},[],3)$. Contrarily, we can compute $\mathcal{X}$ from $\hat{\mathcal{X}}$ using the inverse FFT, i.e., $\mathcal{X}=\mathrm{ifft}(\hat{\mathcal{X}},[],3)$.

The work\cite{kilmer2011factorization} give the first definition of the \textbf{block circulation} operation for a tensor $\mathcal{X} \in \mathbb{R}^{n_1\times n_2\times n_3}:$
\begin{equation*}       
	\mathrm{bcirc}(\mathcal{X}):=\left[                 
	\begin{array}{cccc}   
		X^{(1)}&\hspace{-2ex}X^{(n_3)}&\hspace{-2ex}\cdots&\hspace{-2ex}X^{(2)}\\ 
		X^{(2)}&\hspace{-2ex}X^{(1)}&\hspace{-2ex}\cdots&\hspace{-2ex}X^{(3)}\\
		\vdots&\hspace{-2ex}\vdots&\hspace{-2ex}\ddots&\hspace{-2ex}\vdots\\
		X^{(n_3)}&\hspace{-2ex}X^{(n_3-1)}&\hspace{-2ex}\cdots&\hspace{-2ex}X^{(1)}\\  
	\end{array}
	\right]              
	\in \mathbb{R}^{n_1 n_3\times n_2 n_3}.
\end{equation*}
The \textbf{block diagonalization} matrix of $\mathcal{X}$ is defined as
\begin{equation*}       
	\mathrm{bdiag}(\mathcal{X}):=\left[               
	\begin{array}{cccc}   
		X^{(1)}&\hspace{-2ex} &\hspace{-2ex} &\hspace{-2ex}  \\ 
		 &\hspace{-2ex} X^{(2)}&\hspace{-2ex} &\hspace{-2ex} \\
		 &\hspace{-2ex} &\hspace{-2ex} \ddots&\hspace{-2ex} \\
		 &\hspace{-2ex} &\hspace{-2ex} &\hspace{-2ex} X^{(n_3)}\\  
	\end{array}
	\right]               
	\in \mathbb{R}^{n_1 n_3\times n_2 n_3}.
\end{equation*}
The block circulant matrix can be block diagonalized, i.e.,
	\begin{equation*}
	\mathrm{bdiag}(\hat{\mathcal{X}})=(F_{n_{3}}\otimes I_{n_{1}})\cdot\mathrm{bcirc}(\mathcal{X})\cdot(F_{n_{3}}^{H}\otimes I_{n_{2}}),
	\end{equation*}
where $F_{n_3}\in\mathbb{C}^{n_3 \times n_3} $ is the discrete Fourier transformation matrix, $I_{n}\in \mathbb{R}^{n \times n}$ is an identity matrix, $\otimes$ denotes the Kronecker product. We also define the following operator
	\begin{equation*}
	\mathrm{bvec}(\mathcal{X})=\left[                 
	\begin{array}{c}  
	X^{(1)}\\ 
	X^{(2)}\\
	\vdots\\
	X^{(n_3)}\\  
	\end{array}
	\right],\mathrm{bvfold}(\mathrm{bvec}(\mathcal{X}))=\mathcal{X}.
	\end{equation*}

\subsection{Tensor Preliminaries}
\newtheorem{definition}{Definition}
\newtheorem{theorem}{Theorem}
\begin{definition}[Tucker Decomposition and Tucker Rank]\cite{kolda2009tensor} 
	The Tucker decomposition is a form of higher-order PCA. It decomposes the tensor into a core tensor multiplied by a matrix along each mode. Given a tensor $\mathcal{X}\in \mathbb{R}^{I\times J\times K}$, it can be decomposed as:
	\begin{equation*}
		\mathcal{X} \approx  \mathcal{G} \times_1 A \times_2 B \times_3 C = \sum_{p = 1}^{P}\sum_{q = 1}^{Q}\sum_{r = 1}^{R}g_{pqr}a_p\circ b_q \circ c_r,
	\end{equation*}
	where ''$\times_n$'' denotes mode-n product, ''$\circ$'' denotes the outer product, $A \in \mathbb{R}^{I\times P}$, $B \in \mathbb{R}^{J\times Q}$, $C \in \mathbb{R}^{K\times R}$ are factor matrices and $\mathcal{G}\in \mathbb{R}^{P\times Q \times R}$ is the core tensor.
	
	A Tucker rank (also known as n-rank) of an N-order tensor is defined the rank of unfolding matrix along each mode, $\bm{r} = (r_1,r_2,\dots,r_N)$, where $r_n$, $n=1,2,\dots,N$, denotes the rank of $X_{(n)}$.
\end{definition}
\begin{definition}[Sum of Nuclear Norm (SNN)]\cite{liu2012tensor}
	The definition of tensor SNN is the weighted average of the nuclear norm of all matrices unfolded along each mode. Mathematically, it can be formulated as:
	\begin{equation*}
		\left\Vert \mathcal{X} \right\Vert_\ast := \sum_{i=1}^{N}\alpha_i\left\Vert X_{(i)} \right\Vert_\ast,
	\end{equation*}
	where $\alpha_i$ denotes the weight of matrix unfolded along $i$-th mode.
\end{definition}
%
\begin{definition}[t-product]\cite{kilmer2011factorization}
	The t-product between two 3-order tensors  $\mathcal{X} \in \mathbb{R}^{n_1\times n_2\times n_3}$ and  $\mathcal{Y} \in \mathbb{R}^{n_2\times n_4\times n_3}$ is defined as:
	\begin{equation*}
		\mathcal{Z}=\mathcal{X}\ast\mathcal{Y}:=\mathrm{bvfold}(\mathrm{bcirc(\mathcal{X})\mathrm{bvec}(\mathcal{X}))}.
	\end{equation*}
	Using the above property, the t-product can be written as:
	\begin{equation*}
		\hat{\mathcal{Z}}=\mathrm{bvfold}(\mathrm{bdiag}(\hat{\mathcal{X}})\mathrm{bvec}(\hat{\mathcal{Y}})),
	\end{equation*}
\end{definition}
\begin{definition}[Other Special Tensor]\cite{kilmer2011factorization}	
	The \textbf{identity tensor} $\mathcal{I}\in \mathbb{R}^{n \times n\times n_3}$ is the tensor whose first frontal slice is the $n \times n$ identity matrix, and other frontal slices are all zeros. 
	The  \textbf{orthogonal tensor}  $\mathcal{Q}\in \mathbb{R}^{n \times n\times n_3}$ is satisfies $\mathcal{Q}\ast\mathcal{Q}^{T}=\mathcal{Q}^{T}\ast\mathcal{Q}=\mathcal{I}$.
	A tensor $\mathcal{X}$ is called \textbf{f-diagonal} if each frontal slice $\mathcal{X}^{(i)}$ is a diagonal matrix.
\end{definition}
\begin{theorem}[t-SVD]\cite{kilmer2011factorization}	
	Let $\mathcal{X}\in \mathbb{R}^{n_1 \times n_2\times n_3}$. Then it can be factored as
	\begin{equation*}
		\mathcal{X}=\mathcal{U}\ast\mathcal{S}\ast\mathcal{V}^{H},
	\end{equation*}
	where $\mathcal{U}\in \mathbb{R}^{n_1 \times n_1\times n_3}$, $\mathcal{V}\in \mathbb{R}^{n_2 \times n_2\times n_3}$ are orthogonal, and $\mathcal{S}\in \mathbb{R}^{n_1 \times n_2\times n_3}$ is an f-diagonal tensor. The t-SVD based on t-product, which can be efficiently obtained by computing a series of matrix SVDs in the Fourier domain. 
\end{theorem}
\begin{definition}[Tensor Tubal Rank]\cite{zhang2014novel}	
	The tensor tubal rank denoted as $rank_t(\mathcal{X})$, is defined as the number of non-zero tubes of $\mathcal{S}$, where $\mathcal{S}$ is from the t-SVD of $\mathcal{X}=\mathcal{U}\ast\mathcal{S}\ast\mathcal{V}^{H}$, that is
	\begin{equation*}
		\mathrm{rank}_{t}(A)= \#\left\{ i:\mathcal{S}(i,:,:)\ne0 \right\}.
	\end{equation*}
\end{definition}
\begin{definition}[Tensor Nuclear Norm (TNN)]\cite{zhang2014novel}	
		Let $\mathcal{X}\in \mathbb{R}^{n_1 \times n_2\times n_3}$, the definition of tensor TNN is the sum of singular values of all the frontal slices of $\hat{\mathcal{X}}$, it can be formulated as:
	\begin{equation*}
		\left\Vert \mathcal{X} \right\Vert_\ast := \sum_{i=1}^{n_{3}}\Vert{\hat{X}^{(i)}}\Vert_\ast.
	\end{equation*}
\end{definition}

\subsection{Convolutional Sparse Coding}
Unlike the traditional sparse coding model, CSC replaces the general linear representation with a sum of convolutions between a set of filters $d_i$ (the set also known as a dictionary) and its corresponding feature maps $\Gamma_i$. Besides, the patch-based dictionary restores the same information in the image several times, while CSC integrates these information only once.
One assumes that an image $X \in \mathbb{R}^{N\times M}$ admits a decomposition as $X = \sum_{i=1}^{K}d_i*\Gamma_i$, where $d_i\in \mathbb{R}^{r\times r}$ denotes the filters, $\Gamma_i \in \mathbb{R}^{N\times M}$ is the feature map corresponding to $d_i$ and $*$ is the convolution operator. The most univesal form of CSC is Convolutional Basis Pursuit DeNoising (CBPDN), formulated as:
\begin{equation}
	\arg\min_{\Gamma_i} \quad\frac{1}{2}\left\Vert \sum_{i=1}^{K}d_i*\Gamma_i-X\right\Vert_F^2+\lambda\sum_{i=1}^{K}\left\Vert\Gamma_i\right\Vert_1.
\end{equation}
In contrast to linear representation, convolutional dictionary can acquire shift-invariant features, which improves the efficiency of the model.
\begin{figure*}
	\centering
		\includegraphics[scale=0.9]{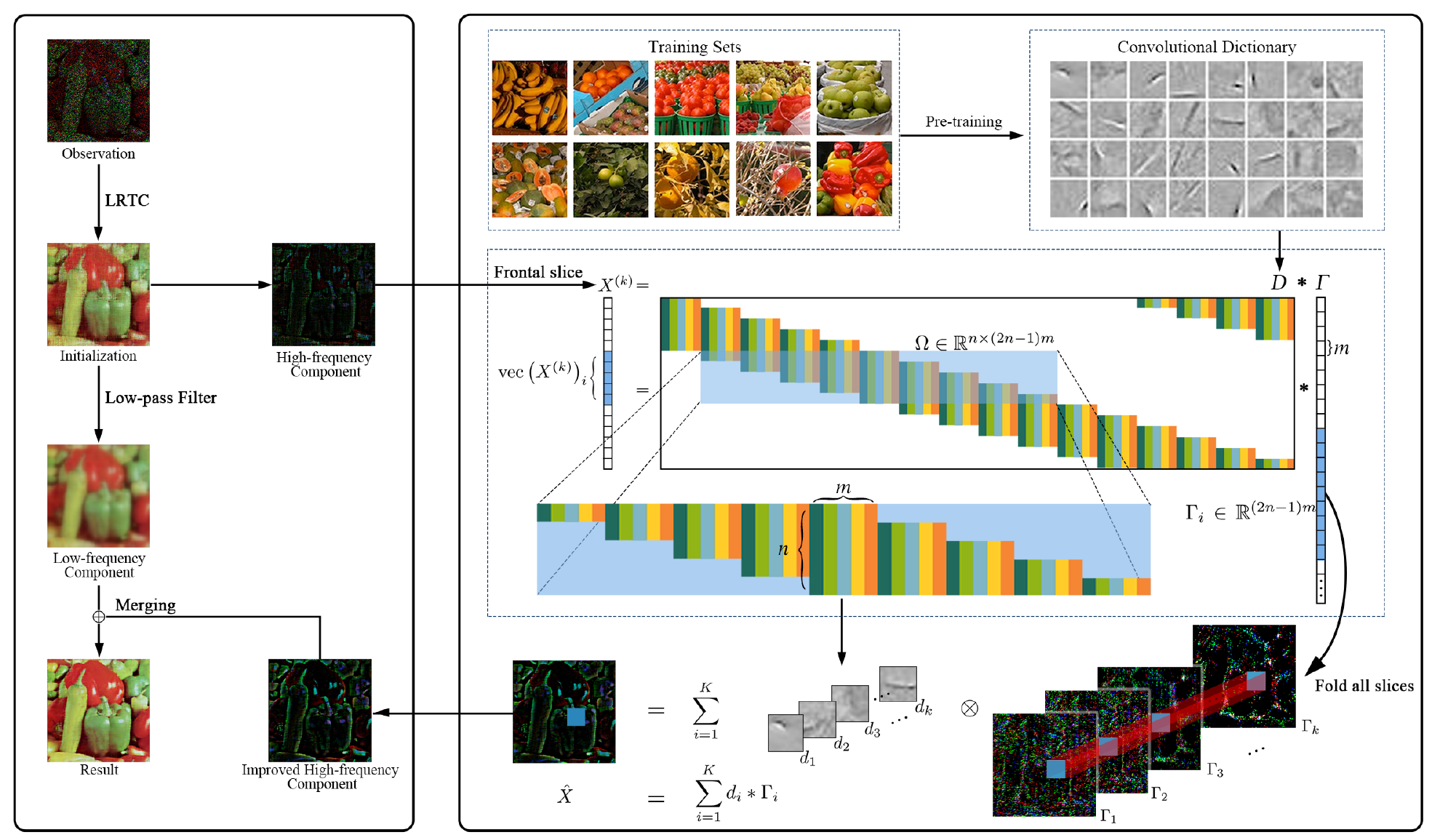}
		\caption{Flowchart of LRTC-CSC models for low-rank tensor completion. The first step is to generate an initial tensor according to observations.  Then a low-pass filter is utilized to divide the initial tensor into two components, a high-frequency and a low-frequency one. Next, the high-frequency component is processed by CSC model, in which the dictionary D (shown in the format of convolutional matrix) is pre-trained by a small-sample dataset.  The stripe dictionary $\Omega$, which is of size $n\times (2n-1)m$, is obtained by extracting the $i$-th patch from the global convolutional dictionary D. The stripe vector $\Gamma_i$, which is the corresponding sparse respresent, contains all coefficients of atoms contributing to $vec(X^{(k)})_i$. At last, by merging the improved high-frequency component and the original low-frequency one, the final result is reconstructed.}
\end{figure*}

%
\section{The Proposed Model and algorithm}
It's imperfect for LRTC models that they can only recover the low-rank component of underlying tensors. In the meanwhile, lots of details are ignored, which derives many composite algorithms with extra priors emerged to improve the missing details. These details often represent the local features or high-frequency component of an image. Previous studies about TV regularization have shown its over-patch (over-smooth) effects. To obtain better details when recovering underlying tensors, 
CSC is introduced to improve the high-frequency component of the underlying data. It's known that the high-frequency component often contains specific detailed information against the low-frequency one. Thus, it's a feasible choice to improve the high-frequency component solely.
By using a pre-trained convolutional dictionary, the corresponding feature maps can be iteratively calculated to approximate the incomplete high-frequency component of tensor data, which is produced by LRTC models. Due to the overcompleteness of the dictionary, the convolution results of obtained feature maps and dictionary would be an improved version of the previous high-frequency component. In this section, we propose to incorporate CSC for different LRTC methods to verify the effectiveness of CSC regularization. Since the low-rank prior of LRTC can be approximated by SNN and TNN, two LRTC-CSC compound methods are developed.


\subsection*{\textbf{Method \hspace{1ex} LRTC-CSC-I}}
As the tratidional typical low-rank approximation method, SNN has gained unexpected achievements in LRTC problems. Even though, its promise is limited by the tensor mode-$n$ flattening operator, which breaks the global structure of the whole tensor. To improve the performance of SNN-based model, we consider to impose a CSC regularization onto the detail component of underlying tensor, mathematically formulated as:
\begin{equation}
	\begin{aligned}
		\arg\min_{\mathcal{X}} \quad \sum_{k=1}^N\alpha_k\left\Vert {X}_{(k)}\right\Vert_\ast+\lambda\Phi(\mathcal{X})
		\quad \mathrm{s.t.}\quad \mathcal{X}_{\Omega}=\mathcal{T}_{\Omega},
	\end{aligned}\label{eqn:lrtc_csc}
\end{equation}
where $\Phi(\mathcal{X})$ denotes the regularization item, i.e. CSC, and $\lambda$ is a trade-off parameter.  

The constraint condition $\mathcal{X}_{\Omega}=\mathcal{T}_{\Omega}$ can be rewritten as a function like:
\begin{equation}
	\min\quad  P(\mathcal{X})=\begin{cases} 
		0,\quad &\mathrm{if} \quad \mathcal{X}_{\Omega}=\mathcal{T}_{\Omega},\\
		\infty,\quad & \mathrm{otherwise}.
	\end{cases}
	\label{eqn:barrier}
\end{equation}

By introducing auxiliary variables  $\{F_{k}\}_{k=1}^{N-1}$ to disentangle the relationship between $\{X_{(k)}\}_{k=1}^{N-1}$ and  $\mathcal{Z} = \mathcal{X}$, the augmented Lagrangian function of (\ref{eqn:lrtc_csc}) can be formulated as:
\begin{equation}
	\begin{aligned}
		\mathcal{L} = &\sum_{k=1}^N\alpha_k\left\Vert F_k\right\Vert_\ast+\sum_{k=1}^{N-1}\frac{\beta_{1_{k}}}{2}\left\Vert F_k - X_{(k)}\right\Vert_F^2\\&+\left\langle\omega_{1_{k}},F_k - X_{(k)}\right\rangle+\frac{\beta_{2}}{2}\left\Vert \mathcal{Z-X}\right\Vert_F^2\\&+\left\langle\omega_{2},\mathcal{Z-X}\right\rangle+P(\mathcal{X})+\lambda\Phi(\mathcal{Z}),
	\end{aligned}
	\label{eqn:lag}
\end{equation}
where $\omega$ is the Lagrangian multiplier and $\beta$  is the penalty parameter.

Under the iterative optimization framework of ADMM \cite{chen2014alternating}, the solution process of (\ref{eqn:lag}) can be concluded by solving three subproblems, i.e. [$F,\mathcal{Z}$] ,$\mathcal{X}$ and $\omega$. 

\subsection{Solving [$F$,$\mathcal{Z}$]-subproblem}
In this part, we are going to solve the [$F$,$\mathcal{Z}$]-subproblem separately, since they are decoupled.

1)  By fixing $\mathcal{X}$ and $\mathcal{Z}$, rewrite the function of $F$-subproblem:
\begin{equation}
	\begin{aligned}
		\arg\min_{F_k} \quad\sum_{k=1}^N\alpha_k\left\Vert F_k\right\Vert_\ast+\sum_{k=1}^{N-1}\frac{\beta_{1_k}}{2}\left\Vert F_k - X_{(k)}+\frac{\omega_{1_k}}{\beta_{1_k}}\right\Vert_F^2.
	\end{aligned}
	\label{eqn:Fpro}
\end{equation}
The iterative closed-form solution of (\ref{eqn:Fpro}) is:
\begin{equation}
	F_k^{i+1}=\mathbf{D}_{\tau_k}(X_{(k)}^{i}-\frac{\omega_{1_k}^i}{\beta_{1_k}})=U\mathrm{diag}(\max(\lambda_l-\tau_k,0))V^T,
	\label{eqn:Fsolution}
\end{equation}
where $\tau_k = \frac{\alpha_k}{\beta_{1_k}}$, $\mathbf{D}_{\tau_k}(X_{(k)})$ denotes the thresholding singular value decomposition (SVD) of $X_{(k)}$ and $\mathrm{diag}(\cdot)$ denotes a diagnal matrix.

2) By fixing $\mathcal{X}$ and $F$, the $\mathcal{Z}$-subproblem is:
\begin{equation}
	\begin{aligned}
		\arg\min_{\mathcal{Z}} \quad \frac{\beta_{2}}{2}\left\Vert \mathcal{Z}-(\mathcal{X}-\frac{\omega_{2}}{\beta_{2}})\right\Vert_F^2+\lambda\Phi(\mathcal{Z}).
	\end{aligned}
	\label{eqn:Zpro}
\end{equation}
Actually as the plug-and-play framework suggested in \cite{zhao2020deep}, the problem
(\ref{eqn:Zpro}) can be regarded as a new denoising problem. $\Phi(\mathcal{Z})$ measures the degree of noise in $\mathcal{Z}$, the smaller, the better.
By treating `` $\mathcal{X}-\frac{\omega_{2}}{\beta_{2}}$ '' as a noisy image and ``$\mathcal{Z}$'' as a clean image, CSC model becomes the denoiser to solve $\mathcal{Z}$-subproblem and outputs a clean image for comparison. 
Thus, we use the noise image `` $\mathcal{X}-\frac{\omega_{2}}{\beta_{2}}$ '' to be the input of CSC model, which can be explicitly formulated as:
\begin{equation}
	\begin{aligned}
		\arg\min_{M_i}\quad	\frac{1}{2}\left\Vert\sum_{i=1}^{K}d_i*M_i-Input\right\Vert_F^2+\Lambda\sum_{i=1}^{K}\left\Vert M_i\right\Vert_1,
		\label{eqn:Zpro2}
	\end{aligned}
\end{equation}
where \textit{Input}$=\mathcal{X}-\frac{\omega_{2}}{\beta_{2}}$ and $\Lambda$ is the hyper-parameter.

In consideration of a fact that only the high-frequency component of \textit{Input }is processed by CSC \cite{bao2019convolutional}, thus, we use a high-pass filter  to acquire the high-frequency component of 
\textit{Input} and in the rest of this section, \textit{Input} represents its high-frequency component. 
Note that \textit{Input} is a $3$-rd tensor while $M$ is a matrix. To keep consistency, \textit{Input} is cut into three frontal slices for three different color channels in color images, repectively. And the following operations are completed at the level of matrices.
At last, the underlying result can be obtained by summing the low-frequency component and recovered high-frequency component.

Because the inaccurate filters will lead to producing new artifacts or structure loss problems, an extra gradient constraint is adopted to suppress the outliers based on the original CSC model. What's more, as suggested in \cite{wohlberg2018convolutional}, the gradient constraint on the feature maps is superior to applying that on the image domain. Thus, we further utilize problem (\ref{eqn:Zpro2}) with gradient constraint on the feature maps, and it can be explicitly formulated as:
\begin{equation}
	\begin{aligned}
		\arg\min_{M_i}\quad	&\frac{1}{2}\left\Vert\sum_{i=1}^{K}d_i*M_i-Input\right\Vert_F^2+\Lambda\sum_{i=1}^{K}\left\Vert M_i\right\Vert_1\\&+\frac{\tau}{2}\sum_{i=1}^{K}\left\Vert \sqrt{\left(g_0*M_i\right)^2+\left(g_1*M_i\right)^2}\right\Vert_F^2\\&
		\label{eqn:Zpro3}
	\end{aligned}
\end{equation}
where $g_0$ and $g_1$ are the filters that compute the gradients along image rows and columns respectively, and $\tau$ is the hyperparameter. By introducing linear operators $G_jM_i=g_j*M_i$, the gradient constraint term of (\ref{eqn:Zpro3}) can be rewritten as:
\begin{equation}
	\frac{\tau}{2}\sum_{i=1}^{K}\left(\left\Vert G_0M_i\right\Vert_F^2+\left\Vert G_1M_i\right\Vert_F^2\right).
	\label{eqn:gradient}
\end{equation}

Further more, considering the conception of block matrix, (\ref{eqn:gradient}) can be simplified as:
\begin{equation}
	\frac{\tau}{2}\left\Vert \Psi_0\mathcal{M}\right\Vert_F^2+\frac{\tau}{2}\left\Vert \Psi_1\mathcal{M}\right\Vert_F^2,
\end{equation}
where 
\begin{equation}
	\begin{aligned}
		\Psi_j=	\left(\begin{array}{cccc} 
			G_j &    0    & \cdots \\ 
			0 &    G_j   & \cdots\\ 
			\vdots & \vdots & \ddots 
		\end{array}\right),
	\end{aligned}
\end{equation}
and $\mathcal{M} = (M_1 M_2 \cdots M_K)^T$.

Let $D_i M_i=d_i*M_i$, $\mathcal{D} = (D_1 D_2 \cdots D_K)$,  the augmented Lagrangian function of (\ref{eqn:Zpro3}) is:
\begin{equation}
	\begin{aligned}
		L_2 = &\frac{1}{2}\left\Vert \mathcal{DM}-Input\right\Vert_F^2+\Lambda\left\Vert \mathcal{B}\right\Vert_1+\left\langle\Theta,\mathcal{M-B}\right\rangle\\&+\frac{\rho}{2}\left\Vert \mathcal{M-B}\right\Vert_F^2	+\frac{\tau}{2}\left\Vert \Psi_0\mathcal{M}\right\Vert_F^2+\frac{\tau}{2}\left\Vert \Psi_1\mathcal{M}\right\Vert_F^2,
	\end{aligned}\label{eqn:lag2}	
\end{equation}
where $\mathcal{B}$ is introduced as an auxiliary variables satisfying $\mathcal{B}=\mathcal{M}$, $\Theta$ is the Lagrangian multiplier and $\rho$  is the penalty parameter.

According to the inexact ADMM framework \cite{hager2019inexact}, the problem (\ref{eqn:lag2}) can be solved by solving a sequence of subproblems.

1) Rewrite the $\mathcal{M}$-subproblem:
\begin{equation}
	\begin{aligned}
		\arg\min_{\mathcal{M}} \quad &\frac{1}{2}\left\Vert \mathcal{DM}-Input\right\Vert_F^2+\frac{\rho}{2}\left\Vert \mathcal{M-B}+C\right\Vert_F^2\\&+\frac{\tau}{2}\left\Vert \Psi_0\mathcal{M}\right\Vert_F^2+\frac{\tau}{2}\left\Vert \Psi_1\mathcal{M}\right\Vert_F^2,
	\end{aligned}\label{eqn:mpro}
\end{equation}
where $C$ denotes the parameter item $C=\frac{\Theta}{\rho}$.
By transforming (\ref{eqn:mpro}) to the Fourier domain, it can be formulated as:
\begin{equation}
	\begin{aligned}
		\arg\min_{\hat{\mathcal{M}}} \quad &\frac{1}{2}\left\Vert \hat{\mathcal{D}}\hat{\mathcal{M}}-\hat{Input}\right\Vert_F^2+\frac{\rho}{2}\left\Vert \hat{\mathcal{M}}-\hat{\mathcal{B}}+\hat{C}\right\Vert_F^2\\&+\frac{\tau}{2}\left\Vert \hat{\Psi}_0\hat{\mathcal{M}}\right\Vert_F^2+\frac{\tau}{2}\left\Vert \hat{\Psi}_1\hat{\mathcal{M}}\right\Vert_F^2,
	\end{aligned}\label{eqn:mprof}
\end{equation}
where $\hat{\mathcal{D}}$, $\hat{\mathcal{M}}$, $\hat{Input}$, $\hat{\mathcal{B}}$, $\hat{C}$, $\hat{\Psi}_0$ and $\hat{\Psi}_1$ denotes the corresponding expressions in the Fourier domain. A closed-form solution of (\ref{eqn:mprof}) is:
\begin{equation}
	\begin{aligned}
		&\left(\hat{\mathcal{D}}^H\hat{\mathcal{D}}+\rho I+\tau\hat{\Psi}_0^H\hat{\Psi}_0+\tau\hat{\Psi}_1^H\hat{\Psi}_1\right)\hat{\mathcal{M}}\\&=\hat{\mathcal{D}}^H\hat{Input}+\rho\left(\hat{\mathcal{B}}-\hat{C}\right),
	\end{aligned}
	\label{eqn:msolution}
\end{equation}
which can be solved by iterated application of the Sherman-Morrison formula \cite{wohlberg2015efficient}. After obtaining $\hat{\mathcal{M}}$, the corresponding $\mathcal{M}$ in spatial domain can be calcaulated by:
\begin{equation}
	\mathcal{M} = \mathrm{ifft}(\hat{\mathcal{M}}),
	\label{eqn:spatialm}
\end{equation}
where $\mathrm{ifft}(\cdot)$ denotes the inverse Fourier transform.

\begin{algorithm}[htbp]
	\caption{LRTC-CSC-I algorithm}
	\label{alg:LRTC-CSC}
	\textbf{Input}: index set $\Omega$, original data $\mathcal{T}$, filters set $\mathcal{D}$
	
	\textbf{Parameters}: $\Lambda, \lambda, \rho, \tau, \alpha_k, \omega_1,  \omega_2, \beta_1,\beta_2$
	
	\textbf{Initialization}: $\mathcal{X}_{\Omega}^0=\mathcal{T}_{\Omega}$, $\mathcal{X}_{\bar{\Omega}}^0=\mathrm{mean}\left(\mathcal{T}\right)$,
	
	\begin{algorithmic}
		\REPEAT 
		\STATE \textbf{F-subproblem}
		\STATE Update $F_k^{i+1}$ by (\ref{eqn:Fsolution})
		\STATE \textbf{$\mathcal{Z}$-subproblem}
		\FOR {$j = 1$ to MaxIter}	
		\STATE Update $\hat{\mathcal{M}}^{j+1}$ by (\ref{eqn:msolution})
		\STATE Update $\mathcal{M}^{j+1}$ by (\ref{eqn:spatialm})
		\STATE Update $\mathcal{B}^{j+1}$ by (\ref{eqn:Bsolution})
		\STATE Update $C^{j+1}$ by (\ref{eqn:Csolution})
		\ENDFOR
		\STATE Update $\mathcal{Z}^{i+1}$ by (\ref{eqn:Zsolution})
		\STATE \textbf{$\mathcal{X}$-subproblem}
		\STATE Update $\mathcal{X}^{i+1}$ by (\ref{eqn:xsolution})
		\STATE \textbf{$\omega$-subproblem}
		\STATE Update $\omega_1^{i+1}$ by (\ref{eqn:omega1solution})
		\STATE Update $\omega_2^{i+1}$ by (\ref{eqn:omega2solution})
		\UNTIL satisfying the stop condition
	\end{algorithmic}
	\textbf{Output}:The recovered tensor $\mathcal{X}$
\end{algorithm}

2) Rewrite the $\mathcal{B}$-subproblem:
\begin{equation}
	\begin{aligned}
		\arg\min_{\mathcal{B}} \quad &\Lambda\left\Vert \mathcal{B}\right\Vert_1+\frac{\rho}{2}\left\Vert \mathcal{M-B}+C\right\Vert_F^2.
	\end{aligned}\label{eqn:bpro}
\end{equation}
By using soft thresholding algorithm, the closed-form solution of (\ref{eqn:bpro}) can be obtained as:
\begin{equation}
	\mathcal{B}^{j+1} = \mathrm{softthresholding}(\mathcal{M}^{j+1}+C^j,\frac{\Lambda}{\rho}).
	\label{eqn:Bsolution}
\end{equation}

3) Update $C$ by fixing $\mathcal{M}$ and $\mathcal{B}$:
\begin{equation}
	C^{j+1} = C^{j} + \Lambda\left(\mathcal{M}^{j+1}-\mathcal{B}^{j+1}\right).
	\label{eqn:Csolution}
\end{equation}

At last, the outputing high-frequency component of $\mathcal{Z}$ can be calculated by:
\begin{equation}
	\mathcal{Z}^{i+1} = \mathcal{D}\mathcal{M}^*,
	\label{eqn:Zsolution}
\end{equation}
where $\mathcal{M}^*$ is the optimal solution of $\mathcal{M}$. And after adding the low-frequency component of \textit{Input}, an entire solution $\mathcal{Z}$ is obtained.

\subsection{Solving $\mathcal{X}$-subproblem}

By fixing $F$ and $\mathcal{Z}$, the $\mathcal{X}$-subproblem can be simplified and rewritten as:
\begin{equation}
	\begin{aligned}
		\arg\min_{\mathcal{X}} \quad &\sum_{k=1}^{N-1}\frac{\beta_{1_k}}{2}\left\Vert F_k - X_{(k)}+\frac{\omega_{1_k}}{\beta_{1_k}}\right\Vert_F^2 +P(\mathcal{X})\\&\frac{\beta_{2}}{2}\left\Vert \mathcal{Z-X}+\frac{\omega_2}{\beta_2}\right\Vert_F^2.
	\end{aligned}
	\label{eqn:xpro}
\end{equation}
The closed-form solution of (\ref{eqn:xpro}) can be obtained by:
\begin{equation}
	\begin{aligned}
		\mathcal{X}^{i+1} =&( \frac{\bar{\beta_1}}{\bar{\beta_1}+\beta_2}\frac{\sum_{k=1}^{N-1}\beta_{1_k}\mathrm{fold}\left(F_k^i+\frac{\omega_{1_k}}{\beta_{1_k}}\right)}{\sum_{k=1}^{N-1}\beta_{1_k}}\\&+
		\frac{\beta_2}{\bar{\beta_1}+\beta_2}(\mathcal{Z}+\frac{\omega_{2}}{\beta_2}))_{\bar{\Omega}}+\mathcal{T},
	\end{aligned}
	\label{eqn:xsolution}
\end{equation}
where $\bar{\Omega}$ is the complement set of $\Omega$.

\subsection{Solving $\omega$-subproblem}
By fixing $\mathcal{X}$ and $F$,  $\mathcal{X}$ and $\mathcal{Z}$, we can update the Lagrangian multiplier $\omega_{1}$ and $\omega_2$, respectively:
\begin{equation}
	\omega_{1_k}^{i+1} = \omega_{1_k}^i+\beta_{1_k}\left(F_k^{i+1}-X_{(k)}^{i+1}\right),
	\label{eqn:omega1solution}
\end{equation}
and
\begin{equation}
	\omega_{2}^{i+1} = \omega_{2}^i+\beta_{2}\left(\mathcal{Z}^{i+1}-\mathcal{X}^{i+1}\right).
	\label{eqn:omega2solution}
\end{equation}

The overall pseudocode of LRTC-CSC-I algorithm is summarized in Algorithm \ref{alg:LRTC-CSC}. 
Actually, our model can degenerate to HaLRTC model \cite{liu2012tensor} by setting the hyperparameter $\lambda=0$.

\subsection*{\textbf{Method \hspace{1ex} LRTC-CSC-II}}
Similar to method I, we use the TNN to approximate the low-rank prior of tensor, instead. The LRTC-CSC-II model is mathematically formulated as:
\begin{equation}
	\begin{aligned}
		\arg\min_{\mathcal{X}} \quad \sum_{k=1}^{n_{3}}\Vert{\hat{X}^{(k)}}\Vert_\ast+\lambda\Phi(\mathcal{X})
		\quad \mathrm{s.t.}\quad \mathcal{X}_{\Omega}=\mathcal{T}_{\Omega}.
	\end{aligned}\label{eqn:lrtc_csc_2}
\end{equation}

The constraint condition can be rewritten as (\ref{eqn:barrier}). By introducing auxiliary variables $\mathcal{F} = \mathcal{X}$ and  $\mathcal{Z} = \mathcal{X}$, the augmented Lagrangian function of (\ref{eqn:lrtc_csc_2}) can be formulated as:
\begin{equation}
	\begin{aligned}
		\mathcal{L} =&\sum_{k=1}^{n_{3}}\left\Vert {\hat{F}^{(k)}}\right\Vert_\ast+\sum_{k=1}^{n_{3}}\frac{\beta_{1}}{2}\left\Vert {F}^{(k)} - {X}^{(k)}\right\Vert_F^2\\&+\left\langle\omega_1^{(k)},{F}^{(k)} - {X}^{(k)}\right\rangle+\frac{\beta_{2}}{2}\left\Vert \mathcal{Z-X}\right\Vert_F^2\\&+\left\langle\omega_{2},\mathcal{Z-X}\right\rangle+P(\mathcal{X})+\lambda\Phi(\mathcal{Z}).
	\end{aligned}
	\label{eqn:lagI}
\end{equation}

In order to distinguish the $k$-th frontal slice and the $k$-th iteration times of a tensor, we use $\mathcal{X}(:,:,k)$ to denote the  $k$-th frontal slice and $\mathcal{X}^k$ for the $k$-th iteration. We divide the problem (\ref{eqn:lagI}) into three subproblems, i.e. [$\mathcal{F},\mathcal{Z}$], $\mathcal{X}$ and $\omega$,  and solve the problem with the same optimization framework of ADMM.

\subsection*{A.\hspace{1ex}Solving [$\mathcal{F}$,$\mathcal{Z}$]-subproblem}
Similarly, we are going to solve the [$\mathcal{F}$,$\mathcal{Z}$]-subproblem separately, since the variables are decoupled.

1)  By fixing $\mathcal{X}$ and $\mathcal{Z}$, rewrite the function of $\mathcal{F}$-subproblem:
\begin{equation}
	\begin{aligned}
		\arg\min_{\mathcal{F}} \quad &\sum_{k=1}^{n_{3}}\left\Vert {\hat{\mathcal{F}}(:,:,k)}\right\Vert_\ast+\\&\sum_{k=1}^{n_{3}}\frac{\beta_1}{2}\left\Vert {\hat{\mathcal{F}}(:,:,k)} -\hat{\mathcal{X}}(:,:,k) + \frac{\omega_{1}(:,:,k)}{\beta_1}\right\Vert_F^2.
	\end{aligned}
	\label{eqn:Fproi}
\end{equation}
The iterative closed-form solution of (\ref{eqn:Fproi}) is:
\begin{equation}
	\begin{aligned}
		\hat{{\mathcal{F}}}^{i+1}(:,:,k)&=\mathbf{D}_{\tau_k}(\hat{{\mathcal{F}}}^{i}(:,:,k)-\frac{\omega_1^i (:,:,k)}{\beta_1})\\&=U\mathrm{diag}(\max(\lambda_l-\tau_k,0))V^T,
	\end{aligned}
	\label{eqn:Fsolutioni}
\end{equation}
where $\tau_k=\frac{1}{\beta_1}$. 
After getting $\hat{{\mathcal{F}}}^{i+1}$, we can get ${\mathcal{F}}^{i+1}$ by inverse Fourier transform
\begin{equation}
	\mathcal{{F}}^{i+1}=\mathrm{ifft}(\mathcal{\hat{F}}^{i+1}).
	\label{eqn:Fsolutionii}
\end{equation}

2) By fixing $\mathcal{X}$ and $F$, the $\mathcal{Z}$-subproblem is:
\begin{equation*}
	\begin{aligned}
		\arg\min_{\mathcal{Z}} \quad \frac{\beta_{2}}{2}\left\Vert \mathcal{Z}-(\mathcal{X}-\frac{\omega_{2}}{\beta_{2}})\right\Vert_F^2+\lambda\Phi(\mathcal{Z}),
	\end{aligned}
	\label{eqn:Zproi}
\end{equation*}
which is the same with Model I. 
The process of solving $\mathcal{Z}$-subproblem has been introduced in detail in Model I, thus we omit them here for readability.

\subsection*{B.\hspace{1ex}Solving $\mathcal{X}$-subproblem}
By fixing $\mathcal{F}$ and $\mathcal{Z}$, the $\mathcal{X}$-subproblem can be simplified and rewritten as:
\begin{equation}
	\begin{aligned}
		\arg\min_{\mathcal{X}}\quad &\sum_{k=1}^{n_{3}}\frac{\beta_1}{2}\left\Vert \hat{\mathcal{F}}(:,:,k) - \hat{\mathcal{X}}(:,:,k)+\frac{\omega_{1} (:,:,k)}{\beta_1}\right\Vert_F^2\\&+\frac{\beta_2}{2}\left\Vert \mathcal{Z}-\mathcal{X}+\frac{\omega_{2}}{\beta_2}\right\Vert_F^2+P(\mathcal{X}).
	\end{aligned}
	\label{eqn:xproi}
\end{equation}
The closed-form solution of (\ref{eqn:xproi}) can be obtained by:
\begin{equation}
	\begin{aligned}
		\mathcal{X}^{i+1} =\left( \frac{\beta_1\mathcal{F}^{i+1}+\beta_2\mathcal{Z}^{i+1}+\omega_{1}^{i}+\omega_{2}^{i}}{\beta_1+\beta_2} \right)_{\bar{\Omega}}+\mathcal{T},
	\end{aligned}
	\label{eqn:xsolutioni}
\end{equation}
where $\bar{\Omega}$ is the complement set of $\Omega$.

\subsection*{C.\hspace{1ex}Solving $\omega$-subproblem}
By fixing $\mathcal{X}$ and $\mathcal{F}$,  $\mathcal{X}$ and $\mathcal{Z}$, we can update the Lagrangian multiplier $\omega_{1}$ and $\omega_2$, respectively:
\begin{equation}
	\begin{cases}
		\omega_{1}^{i+1} = \omega_{1}^i+\beta_{1}\left(\mathcal{F}^{i+1}-\mathcal{X}^{i+1}\right),\\
		\omega_{2}^{i+1} =   \omega_{2}^i+\beta_{2}\left(\mathcal{Z}^{i+1}-\mathcal{X}^{i+1}\right).\\
	\end{cases}
	\label{eqn:omega1solutioni}
\end{equation}

The overall pseudocode of LRTC-CSC-II algorithm is summarized in Algorithm \ref{alg:LRTC-CSC-II}. 
Actually, our model can degenerate to LRTC-TNN model \cite{zhang2014novel} by setting the hyperparameter $\lambda=0$.

\begin{algorithm}[htbp]
	\caption{LRTC-CSC-II algorithm}
	\label{alg:LRTC-CSC-II}
	\textbf{Input}: index set $\Omega$, original data $\mathcal{T}$, filters set $\mathcal{D}$
	
	\textbf{Parameters}: $\Lambda, \lambda, \rho, \tau, \alpha_k, \omega_1,  \omega_2, \beta_1,\beta_2$
	
	\textbf{Initialization}: $\mathcal{X}_{\Omega}^0=\mathcal{T}_{\Omega}$, $\mathcal{X}_{\bar{\Omega}}^0=\mathrm{mean}\left(\mathcal{T}\right)$,
	
	\begin{algorithmic}
		\REPEAT 
		\STATE \textbf{$\mathcal{F}$-subproblem}
		\STATE Update $\mathcal{{F}}^{i+1}$ by (\ref{eqn:Fsolutionii})
		\STATE \textbf{$\mathcal{Z}$-subproblem}
		\FOR {$j = 1$ to MaxIter}	
		\STATE Update $\hat{\mathcal{M}}^{j+1}$ by (\ref{eqn:msolution})
		\STATE Update $\mathcal{M}^{j+1}$ by (\ref{eqn:spatialm})
		\STATE Update $\mathcal{B}^{j+1}$ by (\ref{eqn:Bsolution})
		\STATE Update $C^{j+1}$ by (\ref{eqn:Csolution})
		\ENDFOR
		\STATE Update $\mathcal{Z}^{i+1}$ by (\ref{eqn:Zsolution})
		\STATE \textbf{$\mathcal{X}$-subproblem}
		\STATE Update $\mathcal{X}^{i+1}$ by (\ref{eqn:xsolutioni})
		\STATE \textbf{$\omega$-subproblem}
		\STATE Update $\omega^{i+1}$ by (\ref{eqn:omega1solutioni})
		\UNTIL satisfying the stop condition
	\end{algorithmic}
	\textbf{Output}:The recovered tensor $\mathcal{X}$
\end{algorithm}

\begin{figure}
	\centering
	\includegraphics[width=0.09\textwidth]{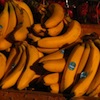}\vspace{1pt}
	\includegraphics[width=0.09\textwidth]{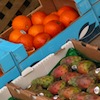}\vspace{1pt}
	\includegraphics[width=0.09\textwidth]{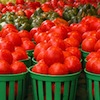}\vspace{1pt}
	\includegraphics[width=0.09\textwidth]{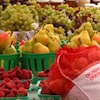}\vspace{1pt}
	\includegraphics[width=0.09\textwidth]{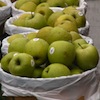}\vspace{1pt}
	\includegraphics[width=0.09\textwidth]{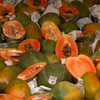}
	\includegraphics[width=0.09\textwidth]{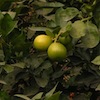}
	\includegraphics[width=0.09\textwidth]{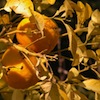}
	\includegraphics[width=0.09\textwidth]{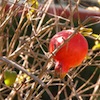}
	\includegraphics[width=0.09\textwidth]{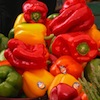}
	\caption{The 10 fruit images for training.}
	\label{fig:dataset}
\end{figure}
\begin{figure}
	\centering
	\includegraphics[width=0.05\textwidth]{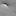}\vspace{0.5pt}
	\includegraphics[width=0.05\textwidth]{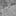}\vspace{0.5pt}
	\includegraphics[width=0.05\textwidth]{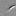}\vspace{0.5pt}
	\includegraphics[width=0.05\textwidth]{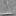}\vspace{0.5pt}
	\includegraphics[width=0.05\textwidth]{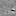}\vspace{0.5pt}
	\includegraphics[width=0.05\textwidth]{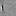}\vspace{0.5pt}
	\includegraphics[width=0.05\textwidth]{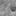}\vspace{0.5pt}
	\includegraphics[width=0.05\textwidth]{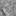}\vspace{0.5pt}
	\includegraphics[width=0.05\textwidth]{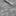}\vspace{0.5pt}
	\includegraphics[width=0.05\textwidth]{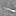}\vspace{0.5pt}
	\includegraphics[width=0.05\textwidth]{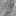}\vspace{0.5pt}
	\includegraphics[width=0.05\textwidth]{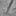}\vspace{0.5pt}
	\includegraphics[width=0.05\textwidth]{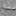}\vspace{0.5pt}
	\includegraphics[width=0.05\textwidth]{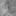}\vspace{0.5pt}
	\includegraphics[width=0.05\textwidth]{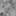}\vspace{0.5pt}
	\includegraphics[width=0.05\textwidth]{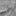}\vspace{0.5pt}
	\includegraphics[width=0.05\textwidth]{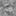}\vspace{0.5pt}
	\includegraphics[width=0.05\textwidth]{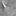}\vspace{0.5pt}
	\includegraphics[width=0.05\textwidth]{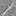}\vspace{0.5pt}
	\includegraphics[width=0.05\textwidth]{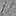}\vspace{0.5pt}
	\includegraphics[width=0.05\textwidth]{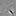}\vspace{0.5pt}
	\includegraphics[width=0.05\textwidth]{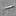}\vspace{0.5pt}
	\includegraphics[width=0.05\textwidth]{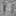}\vspace{0.5pt}
	\includegraphics[width=0.05\textwidth]{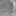}\vspace{0.5pt}
	\includegraphics[width=0.05\textwidth]{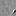}
	\includegraphics[width=0.05\textwidth]{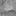}
	\includegraphics[width=0.05\textwidth]{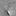}
	\includegraphics[width=0.05\textwidth]{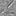}
	\includegraphics[width=0.05\textwidth]{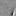}
	\includegraphics[width=0.05\textwidth]{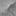}
	\includegraphics[width=0.05\textwidth]{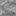}
	\includegraphics[width=0.05\textwidth]{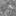}
	\caption{The dictionary of 32 filters trained with fruit dataset.}
	\label{fig:dictionary}
\end{figure}

\begin{figure}[tp]
	\centering
	\includegraphics[width=0.25\textwidth]{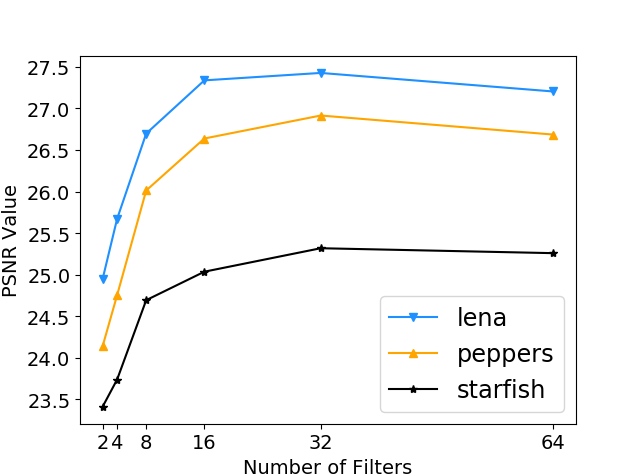}\hspace{-10pt}
	\includegraphics[width=0.25\textwidth]{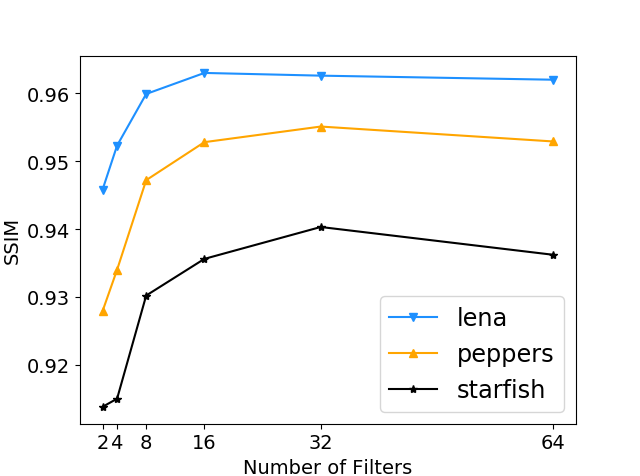}
	\caption{The PSNR and SSIM values of color images recovered by LRTC-CSC when different number of filters are used in a dictionary for the sampling rate 20\%.}
	\label{fig:filter_num}
\end{figure}

\begin{figure*}[htbp]
	\centering
	\subfigure[]{
		\begin{minipage}[b]{0.095\textwidth}
			\centering
			{\footnotesize Original image}\vspace{2pt} 
			\includegraphics[width=\textwidth]{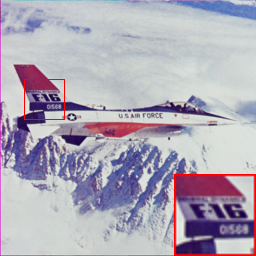}\vspace{2pt}
			\includegraphics[width=\textwidth]{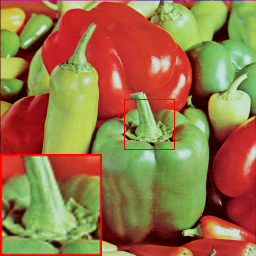}\vspace{2pt}
			\includegraphics[width=\textwidth]{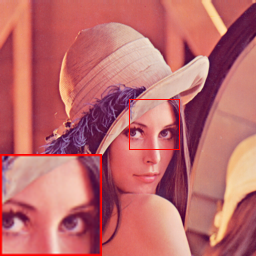}\vspace{2pt}
			\includegraphics[width=\textwidth]{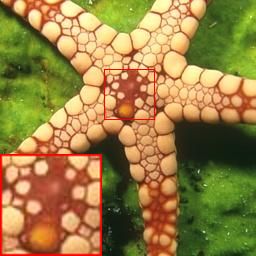}\vspace{2pt}
			\includegraphics[width=\textwidth]{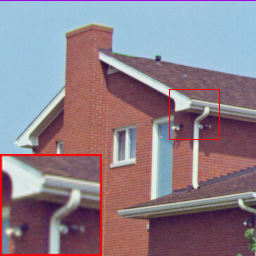}\vspace{2pt}
			\includegraphics[width=\textwidth]{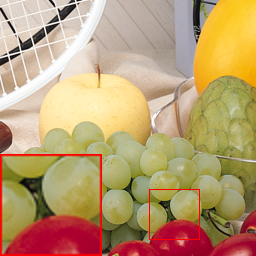}\vspace{2pt}
			\includegraphics[width=\textwidth]{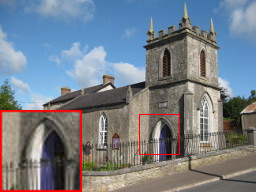}\vspace{2pt}
			\includegraphics[width=\textwidth]{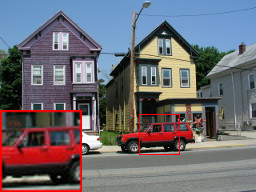}\vspace{2pt}
			
		\end{minipage}
	}
	\hspace{-2.5ex}
	\subfigure[]{
		\begin{minipage}[b]{0.095\textwidth}
			\centering
			\footnotesize Observed\vspace{2pt}
			\includegraphics[width=\textwidth]{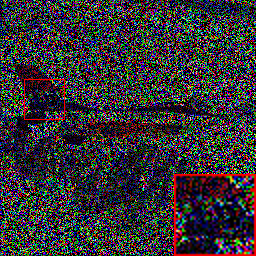}\vspace{2pt}
			\includegraphics[width=\textwidth]{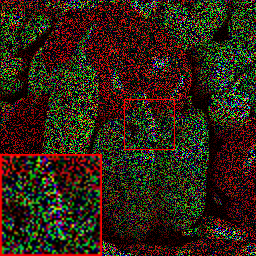}\vspace{2pt}
			\includegraphics[width=\textwidth]{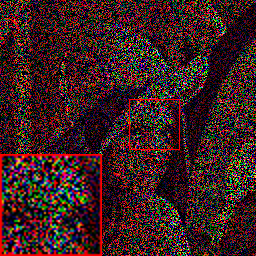}\vspace{2pt}
			\includegraphics[width=\textwidth]{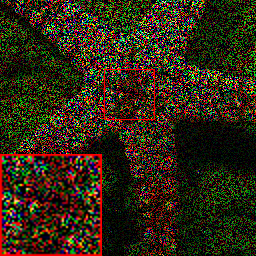}\vspace{2pt}
			\includegraphics[width=\textwidth]{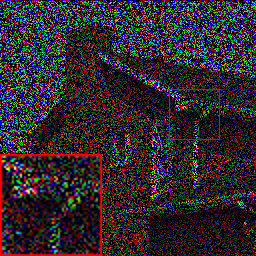}\vspace{2pt}
			\includegraphics[width=\textwidth]{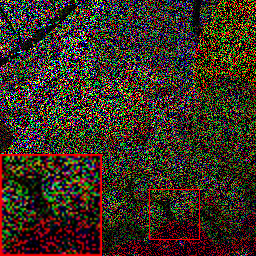}\vspace{2pt}
			\includegraphics[width=\textwidth]{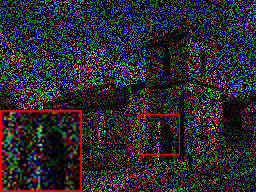}\vspace{2pt}
			\includegraphics[width=\textwidth]{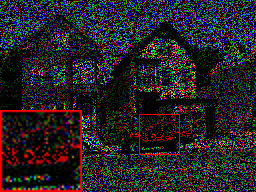}\vspace{2pt}
			
		\end{minipage}
	}
	\hspace{-2.5ex}
	\subfigure[]{
		\begin{minipage}[b]{0.095\textwidth}
			\centering
			\footnotesize SNN\vspace{2pt}
			\includegraphics[width=\textwidth]{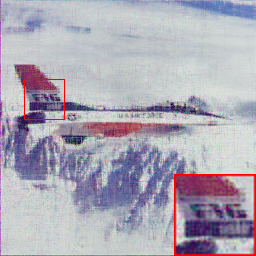}\vspace{2pt}
			\includegraphics[width=\textwidth]{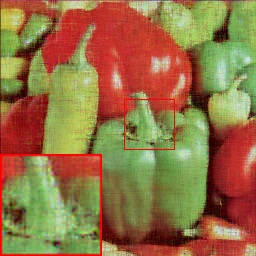}\vspace{2pt}
			\includegraphics[width=\textwidth]{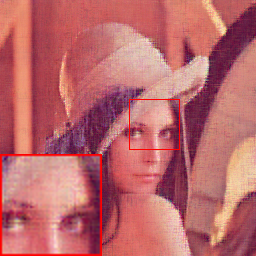}\vspace{2pt}
			\includegraphics[width=\textwidth]{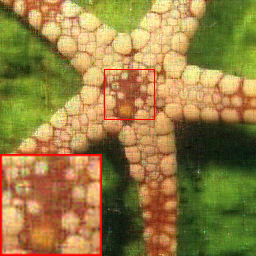}\vspace{2pt}
			\includegraphics[width=\textwidth]{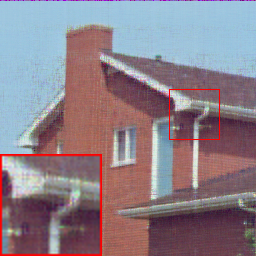}\vspace{2pt}
			\includegraphics[width=\textwidth]{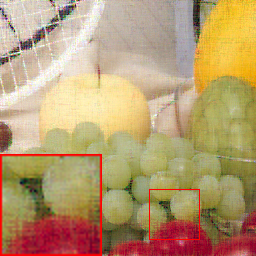}\vspace{2pt}	
			\includegraphics[width=\textwidth]{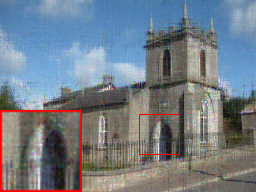}\vspace{2pt}
			\includegraphics[width=\textwidth]{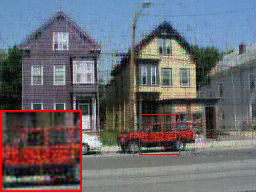}\vspace{2pt}
			
		\end{minipage}
	}
	\hspace{-2.5ex}
	\subfigure[]{
		\begin{minipage}[b]{0.095\textwidth}
			\centering
			\footnotesize TNN\vspace{2pt}
			\includegraphics[width=\textwidth]{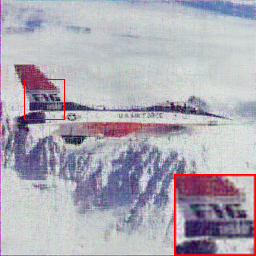}\vspace{2pt}
			\includegraphics[width=\textwidth]{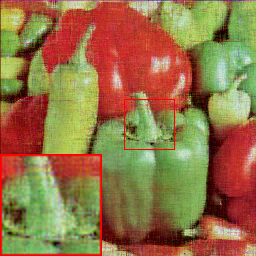}\vspace{2pt}
			\includegraphics[width=\textwidth]{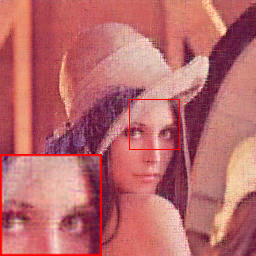}\vspace{2pt}
			\includegraphics[width=\textwidth]{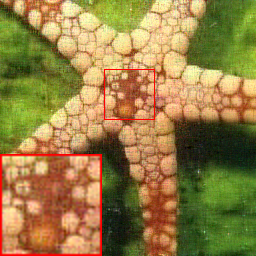}\vspace{2pt}
			\includegraphics[width=\textwidth]{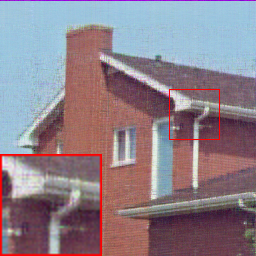}\vspace{2pt}
			\includegraphics[width=\textwidth]{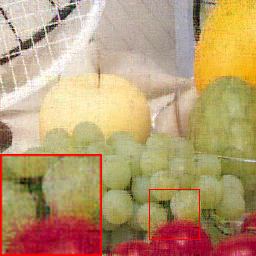}\vspace{2pt}
			\includegraphics[width=\textwidth]{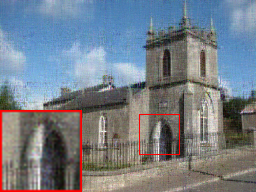}\vspace{2pt}
			\includegraphics[width=\textwidth]{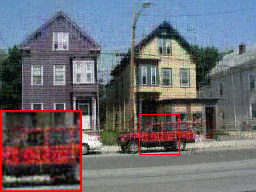}\vspace{2pt}
			
		\end{minipage}
	}
	\hspace{-2.5ex}
	\subfigure[]{
		\begin{minipage}[b]{0.095\textwidth}
			\centering
			\footnotesize TV\vspace{2pt}
			\includegraphics[width=\textwidth]{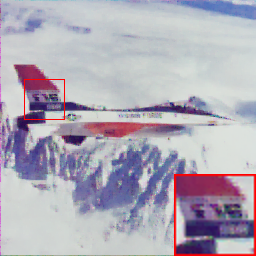}\vspace{2pt}
			\includegraphics[width=\textwidth]{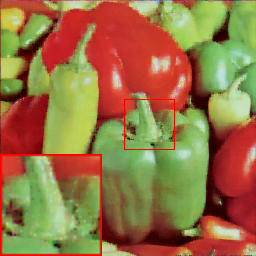}\vspace{2pt}
			\includegraphics[width=\textwidth]{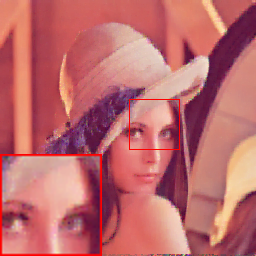}\vspace{2pt}
			\includegraphics[width=\textwidth]{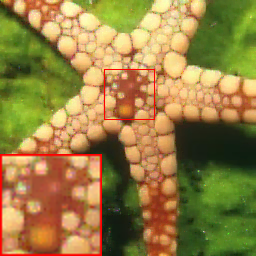}\vspace{2pt}
			\includegraphics[width=\textwidth]{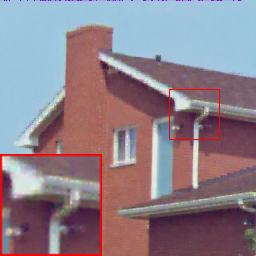}\vspace{2pt}
			\includegraphics[width=\textwidth]{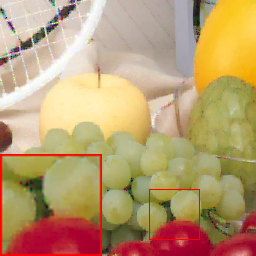}\vspace{2pt}
			\includegraphics[width=\textwidth]{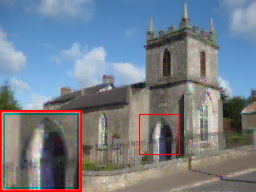}\vspace{2pt}
			\includegraphics[width=\textwidth]{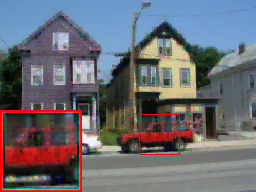}\vspace{2pt}
			
		\end{minipage}
	}
	\hspace{-2.5ex}
		\subfigure[]{
		\begin{minipage}[b]{0.095\textwidth}
			\centering
			\footnotesize Framelet\vspace{2pt}
			\includegraphics[width=\textwidth]{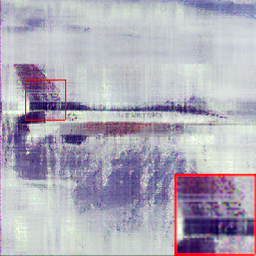}\vspace{2pt}
			\includegraphics[width=\textwidth]{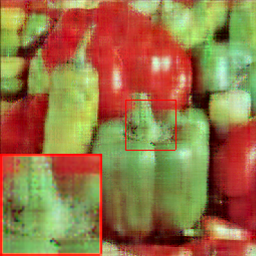}\vspace{2pt}
			\includegraphics[width=\textwidth]{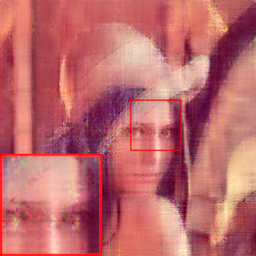}\vspace{2pt}
			\includegraphics[width=\textwidth]{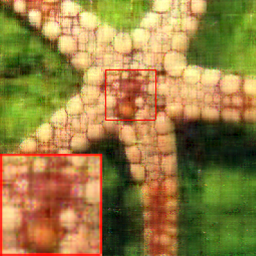}\vspace{2pt}
			\includegraphics[width=\textwidth]{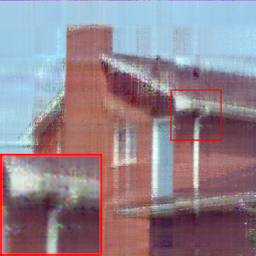}\vspace{2pt}
			\includegraphics[width=\textwidth]{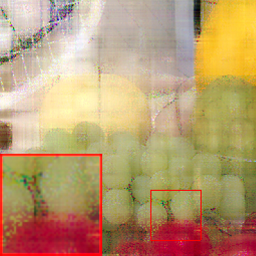}\vspace{2pt}
			\includegraphics[width=\textwidth]{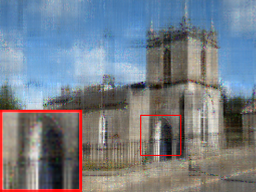}\vspace{2pt}
			\includegraphics[width=\textwidth]{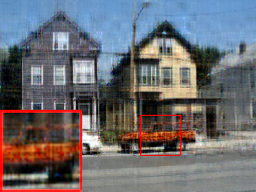}\vspace{2pt}
			
		\end{minipage}
	}
	\hspace{-2.5ex}
	\subfigure[]{
		\begin{minipage}[b]{0.095\textwidth}
			\centering
			\footnotesize 3DTV\vspace{2pt}
			\includegraphics[width=\textwidth]{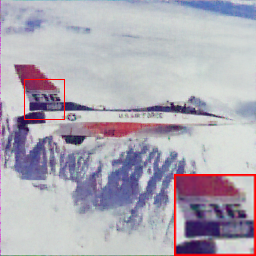}\vspace{2pt}
			\includegraphics[width=\textwidth]{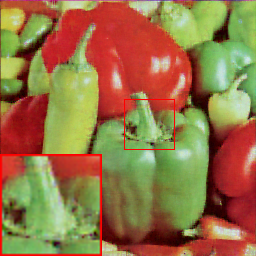}\vspace{2pt}
			\includegraphics[width=\textwidth]{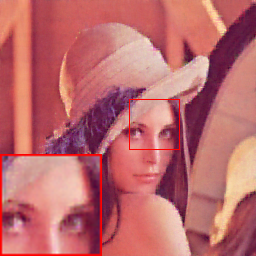}\vspace{2pt}
			\includegraphics[width=\textwidth]{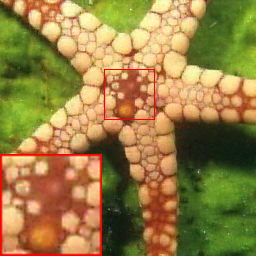}\vspace{2pt}
			\includegraphics[width=\textwidth]{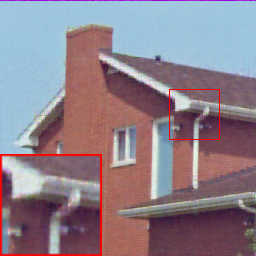}\vspace{2pt}
			\includegraphics[width=\textwidth]{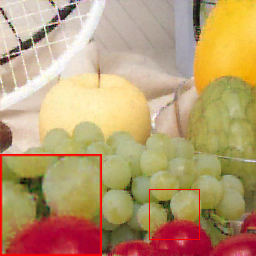}\vspace{2pt}
			\includegraphics[width=\textwidth]{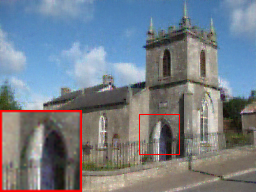}\vspace{2pt}
			\includegraphics[width=\textwidth]{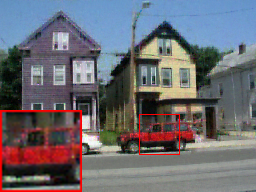}\vspace{2pt}
			
		\end{minipage}
	}
	\hspace{-2.5ex}
	\subfigure[]{
		\begin{minipage}[b]{0.095\textwidth}
			\centering
			\footnotesize Framelet-TV\vspace{2pt}
			\includegraphics[width=\textwidth]{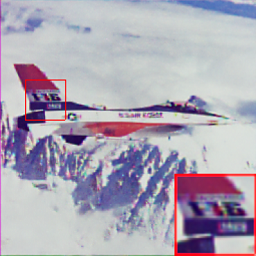}\vspace{2pt}
			\includegraphics[width=\textwidth]{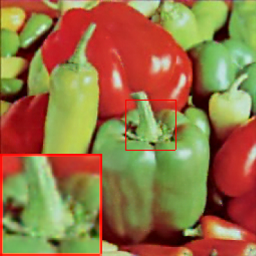}\vspace{2pt}
			\includegraphics[width=\textwidth]{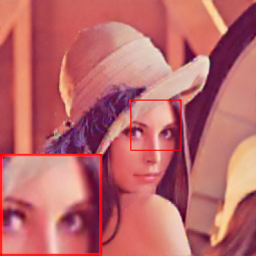}\vspace{2pt}
			\includegraphics[width=\textwidth]{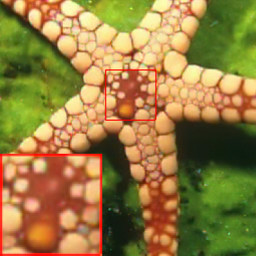}\vspace{2pt}
			\includegraphics[width=\textwidth]{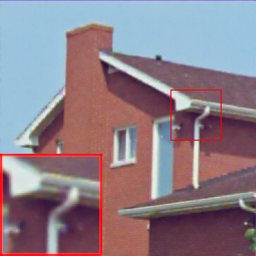}\vspace{2pt}
			\includegraphics[width=\textwidth]{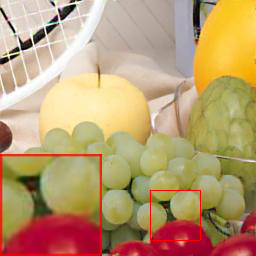}\vspace{2pt}
			\includegraphics[width=\textwidth]{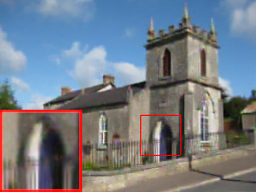}\vspace{2pt}
			\includegraphics[width=\textwidth]{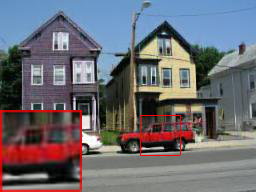}\vspace{2pt}
			
		\end{minipage}
	}
	\hspace{-2.5ex}
		\subfigure[]{
		\begin{minipage}[b]{0.095\textwidth}
			\centering
			\footnotesize CSC-I\vspace{2pt}
			\includegraphics[width=\textwidth]{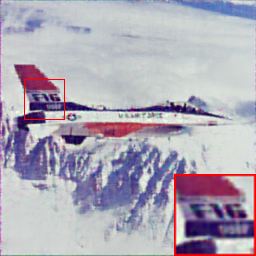}\vspace{2pt}
			\includegraphics[width=\textwidth]{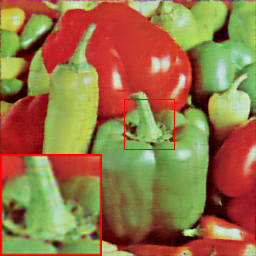}\vspace{2pt}
			\includegraphics[width=\textwidth]{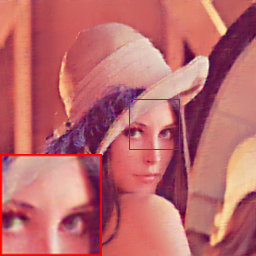}\vspace{2pt}
			\includegraphics[width=\textwidth]{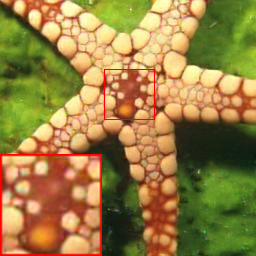}\vspace{2pt}
			\includegraphics[width=\textwidth]{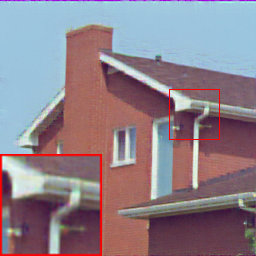}\vspace{2pt}
			\includegraphics[width=\textwidth]{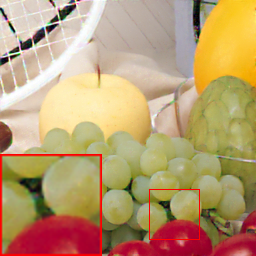}\vspace{2pt}
			\includegraphics[width=\textwidth]{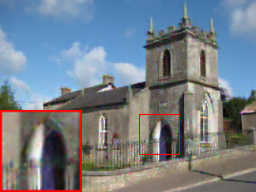}\vspace{2pt}
			\includegraphics[width=\textwidth]{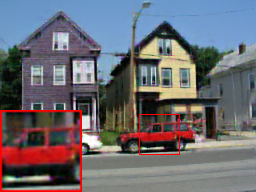}\vspace{2pt}
			
		\end{minipage}
	}
	\hspace{-2.5ex}
	\subfigure[]{
	\begin{minipage}[b]{0.095\textwidth}
		\centering
		\footnotesize CSC-II\vspace{2pt}
		\includegraphics[width=\textwidth]{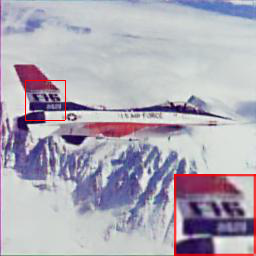}\vspace{2pt}
		\includegraphics[width=\textwidth]{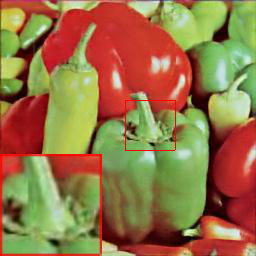}\vspace{2pt}
		\includegraphics[width=\textwidth]{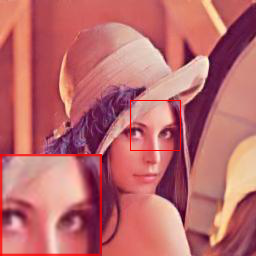}\vspace{2pt}
		\includegraphics[width=\textwidth]{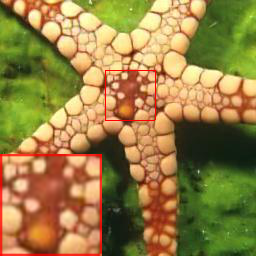}\vspace{2pt}
		\includegraphics[width=\textwidth]{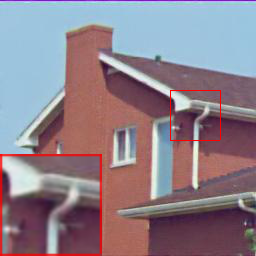}\vspace{2pt}
		\includegraphics[width=\textwidth]{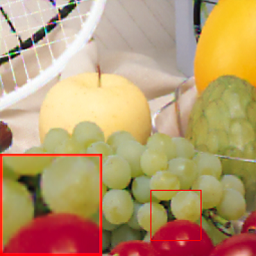}\vspace{2pt}
		\includegraphics[width=\textwidth]{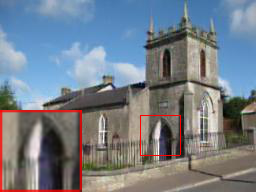}\vspace{2pt}
		\includegraphics[width=\textwidth]{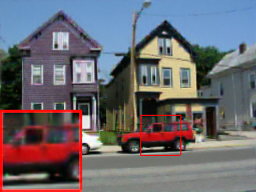}\vspace{2pt}
		
	\end{minipage}
}
	\hspace{-2.5ex}
	\caption{The recovered color images by different algorithms for the sampling rate $30\%$, respectively. From (a) to (j): Original image, Observed image, recovered images by SNN, TNN, TV, Framelet, 3DTV, Framelet-TV, CSC-I and CSC-II, respectively.}
	\label{fig:colorimage}
\end{figure*}

\section{ Numerical Experiments}
In this section, we verified our LRTC-CSC models with popular color image datasets, such as \textit{Peppers}, \textit{Lena}, \textit{Starfish} and so on. Besides, MRI data and video are also leveraged to our benchmark, which shows the generalization of LRTC-CSC model. Meanwhile, several state-of-the-art LRTC methods are compared in different situations:
\begin{enumerate}
	\item HaLRTC \cite{liu2012tensor}: It adopted SNN to approximate the low-rank prior of underlying tensors without any regularizations in the objective function.
	\item LRTC-TNN \cite{zhang2014novel}: It used TNN to describe the low-rank prior of underlying tensors without any regularizations in the objective function.
	\item LRTC-TV \cite{li2017low}: It combined the Tucker rank-based SNN fidelity item and TV regularization to approximate the global low-rank structure and local smoothness, respectively.
	\item  TNN-3DTV \cite{jiang2018anisotropic}: It combined TNN and anisotopic TV regularization, inorder to extract the intrinsic structures of visual data and exploit the local smooth and piecewise priors, simultaneously.
	\item MF-Framelet \cite{jiang2018matrix}: It leveraged matrix factorization-based SNN to capture the global structure of underlying tensor with the Framelet regularization.
	\item Framelet-TV \cite{yang2020tensor}: It combined the TT rank-based SNN and the hybrid regularization of Framelet and TV, aiming at characterizing the global TT low-rankness, capturing the abundant details and enhancing the temporal smoothness of the tensor, respectively.
\end{enumerate}

To measure the recovering performance of various models, both the Peak Signal to Noise Ratio (PSNR) and Structural SIMilarity Index (SSIM) are used. In consideration of the multiband structure of tensor data, we adopt the mean value of all bands as the evaluation metric. 

All experiments are performed in MATLAB R2020a in Linux with an Inter(R) Core(TM) i7-9800X CPU at 3.80 GHz and 64GB RAM. To accelerate the running speed, we performed the CSC phase with a graphics processing GeForce RTX 2080 Ti.

\begin{table*}[tp]  
	
	\centering  
	\begin{threeparttable}  
		\caption{PSNR results of recovered color images by different algorithms. The \textbf{best} and \uline{second} best values are highlighted in bold and underline, respectively. (MR is short for Missing Ratio)}  
		\label{tab:psnr}  
		\begin{tabular}{cccccccccc}  
			\toprule  
			\multirow{3}{*}{Image}& \multirow{3}{*}{MR}&   
			\multicolumn{8}{c}{PSNR}\cr
			\cmidrule(lr){3-10} 
			&&SNN&TNN&TV&Framelet&3DTV&Framelet-TV&CSC-I&CSC-II\cr  
			\midrule  
			\multirow{3}{*}{\shortstack{\textit{Peppers}\\$256\times256\times3$}}
			&70\% &24.03&23.54&26.87&22.66&27.45&30.43&\uline{30.84}&\textbf{31.72}\cr  
			&80\% &20.99&20.50&24.52&21.63&24.82&28.02&\uline{28.20}&\textbf{28.33}\cr  
			&90\% &16.92&16.75&20.24&18.52&20.61&\uline{23.96}&23.52&\textbf{24.08}\cr
			\midrule  
			\multirow{3}{*}{\shortstack{\textit{Starfish}\\$256\times256\times3$}}
			&70\% &23.68&24.41&26.18&22.36&27.27&27.88&\uline{28.50}&\textbf{29.78}\cr  
			&80\% &20.90&21.17&23.55&21.41&24.47&25.18&\uline{25.86}&\textbf{26.75}\cr  
			&90\% &17.29&17.54&20.19&17.90&20.71&21.15&\uline{22.20}&\textbf{22.63}\cr
			\midrule  
			\multirow{3}{*}{\shortstack{\textit{Lena}\\$256\times256\times3$}}
			&70\% &25.59&25.86&27.69&22.75&28.72&29.33&\uline{30.73}&\textbf{32.03}\cr  
			&80\% &22.82&23.03&25.84&21.87&26.26&26.99&\uline{28.13}&\textbf{29.00}\cr  
			&90\% &19.26&19.47&22.29&20.59&22.80&23.94&\uline{24.58}&\textbf{25.34}\cr
			\midrule  
			\multirow{3}{*}{\shortstack{\textit{Fruits}\\$256\times256\times3$}}
			&70\% &24.08&24.30&26.81&21.15&27.21&28.52&\uline{29.44}&\textbf{30.36}\cr  
			&80\% &21.65&21.79&24.89&20.41&25.07&26.43&\uline{26.70}&\textbf{27.59}\cr  
			&90\% &18.22&18.40&21.69&19.18&21.69&23.54&\uline{23.95}&\textbf{24.26}\cr
			\midrule  
			\multirow{3}{*}{\shortstack{\textit{House}\\$256\times256\times3$}}
			&70\% &27.30&27.83&28.45&23.09&30.62&31.82&\uline{32.26}&\textbf{33.00}\cr  
			&80\% &24.22&24.57&26.53&21.81&27.76&29.07&\uline{29.81}&\textbf{30.72}\cr  
			&90\% &20.60&20.62&22.50&20.53&23.78&25.29&\uline{25.62}&\textbf{26.49}\cr
			\midrule  
			\multirow{3}{*}{\shortstack{\textit{Airplane}\\$256\times256\times3$}}
			&70\% &25.21&25.49&26.55&21.03&27.68&26.63&\uline{28.42}&\textbf{30.05}\cr  
			&80\% &22.70&22.91&24.13&20.05&25.27&24.42&\uline{26.12}&\textbf{27.37}\cr  
			&90\% &19.54&19.76&20.96&18.94&22.02&21.84&\uline{22.98}&\textbf{23.92}\cr
			\midrule  
			\multirow{3}{*}{\shortstack{\textit{Church}\\$256\times192\times3$}}
			&70\% &24.70&25.45&25.83&22.92&27.04&\uline{28.32}&27.46&\textbf{28.68}\cr  
			&80\% &22.30&22.84&23.90&21.90&24.81&\uline{26.02}&25.54&\textbf{26.49}\cr  
			&90\% &18.79&19.33&20.82&20.35&21.69&\uline{22.55}&22.41&\textbf{23.21}\cr
			\midrule  
			\multirow{3}{*}{\shortstack{\textit{Cars}\\$256\times192\times3$}}
			&70\% &22.76&23.66&24.18&22.60&25.30&\uline{27.18}&26.31&\textbf{27.41}\cr 
			&80\% &20.15&20.86&21.98&21.23&22.78&\uline{24.62}&24.07&\textbf{24.80}\cr
			&90\% &16.84&17.60&18.34&16.24&19.52&\uline{21.02}&20.76&\textbf{21.26}\cr     
			\bottomrule  
		\end{tabular}  
	\end{threeparttable}  
\end{table*} 

\begin{table*}[tp]  
	
	\centering  
	\begin{threeparttable}  
		\caption{SSIM results of recovered color images by different algorithms. The \textbf{best} and \uline{second} best values are highlighted in bold and underline, respectively. (MR is short for Missing Ratio)}  
		\label{tab:ssim}  
		\begin{tabular}{cccccccccc}  
			\toprule  
			\multirow{3}{*}{Image}& \multirow{3}{*}{MR}&   
			\multicolumn{8}{c}{SSIM}\cr
			\cmidrule(lr){3-10} 
			&&SNN&TNN&TV&Framelet&3DTV&Framelet-TV&CSC-I&CSC-II\cr  
			\midrule  
		\multirow{3}{*}{\shortstack{\textit{Peppers}\\$256\times256\times3$}}
		&70\% &0.9339&0.9233&0.9703&0.9147&0.9687&0.9759&\uline{0.9855}&\textbf{0.9865}\cr  
		&80\% &0.8827&0.8580&0.9501&0.8963&0.9450&\uline{0.9746}&0.9734&\textbf{0.9751}\cr  
		&90\% &0.7588&0.7099&0.8906&0.8168&0.8713&\uline{0.9362}&0.9307&\textbf{0.9406}\cr
		\midrule  
		\multirow{3}{*}{\shortstack{\textit{Starfish}\\$256\times256\times3$}}
		&70\% &0.9074&0.8986&0.9589&0.8873&0.9574&0.9696&\uline{0.9705}&\textbf{0.9768}\cr  
		&80\% &0.8477&0.8255&0.9272&0.8605&0.9250&0.9504&\uline{0.9512}&\textbf{0.9587}\cr  
		&90\% &0.7320&0.6801&0.8801&0.7323&0.8482&0.9007&\uline{0.9013}&\textbf{0.9074}\cr
		\midrule  
		\multirow{3}{*}{\shortstack{\textit{Lena}\\$256\times256\times3$}}
		&70\% &0.9477&0.9460&0.9668&0.9156&0.9722&0.9778&\uline{0.9814}&\textbf{0.9853}\cr  
		&80\% &0.9128&0.9041&0.9511&0.8992&0.9536&0.9642&\uline{0.9687}&\textbf{0.9728}\cr  
		&90\% &0.8450&0.8128&0.9102&0.8684&0.9104&0.9376&\uline{0.9398}&\textbf{0.9436}\cr
		\midrule  
		\multirow{3}{*}{\shortstack{\textit{Fruits}\\$256\times256\times3$}}
		&70\% &0.8747&0.8643&0.9475&0.8328&0.9374&0.9619&\uline{0.9645}&\textbf{0.9729}\cr  
		&80\% &0.8174&0.7962&0.9213&0.8053&0.9055&0.9409&\uline{0.9433}&\textbf{0.9495}\cr  
		&90\% &0.7164&0.6575&0.8607&0.7499&0.8294&\uline{0.8962}&0.8920&\textbf{0.9014}\cr
		\midrule  
		\multirow{3}{*}{\shortstack{\textit{House}\\$256\times256\times3$}}
		&70\% &0.9233&0.9149&0.9562&0.8593&0.9571&0.9713&\uline{0.9717}&\textbf{0.9745}\cr  
		&80\% &0.8610&0.8406&0.9317&0.8242&0.9244&0.9534&\uline{0.9550}&\textbf{0.9601}\cr  
		&90\% &0.7276&0.6642&0.8611&0.7658&0.8326&0.9009&\uline{0.9062}&\textbf{0.9156}\cr
		\midrule  
		\multirow{3}{*}{\shortstack{\textit{Airplane}\\$256\times256\times3$}}
		&70\% &0.6588&0.6215&\uline{0.9132}&0.6761&0.8104&0.9078&0.9115&\textbf{0.9458}\cr  
		&80\% &0.5407&0.4907&0.8580&0.5971&0.7131&0.8613&\uline{0.8630}&\textbf{0.9103}\cr  
		&90\% &0.3668&0.2914&0.7458&0.5090&0.5192&0.7610&\uline{0.7678}&\textbf{0.8184}\cr
		\midrule  
		\multirow{3}{*}{\shortstack{\textit{Church}\\$256\times192\times3$}}
		&70\% &0.8327&0.8485&0.8798&0.7922&0.8933&\uline{0.9242}&0.9130&\textbf{0.9313}\cr  
		&80\% &0.7367&0.7497&0.8202&0.7491&0.8271&\uline{0.8828}&0.8693&\textbf{0.8923}\cr  
		&90\% &0.5420&0.5555&0.6959&0.6629&0.6888&\uline{0.7835}&0.7728&\textbf{0.7918}\cr
		\midrule  
		\multirow{3}{*}{\shortstack{\textit{Cars}\\$256\times192\times3$}}
		&70\% &0.7817&0.8021&0.8664&0.7889&0.8618&\uline{0.9154}&0.9058&\textbf{0.9218}\cr 
		&80\% &0.6628&0.6780&0.7902&0.7320&0.7717&\uline{0.8216}&0.8541&\textbf{0.8672}\cr
		&90\% &0.4378&0.4513&0.6106&0.5411&0.5795&\uline{0.7354}&0.7309&\textbf{0.7659}\cr     
		\bottomrule  
		\end{tabular}  
	\end{threeparttable}  
\end{table*}

\subsection{Parameters}
The convolutional filters are pre-trained with the popular fruit dataset \cite{zeiler2010deconvolutional} first in color image experiments as shown in Fig. \ref{fig:dataset}. To make sure the best number of filters, we trained several dictionaries of different number of filters with a fruit dataset. And then these dictionaries were used to our benchmark for the sampling rate 20\%. The experimental results are shown in Fig. \ref{fig:filter_num}. The best performance corresponds to the best number of filters. All the results in the left figure of Fig. \ref{fig:filter_num} indicated that the proposed model would gain the best performance when a dictionary consists of 32 filters. Except for the \textit{Lena} image, the other results in the right figure of Fig. \ref{fig:filter_num} also show the superiority when the filter number is 32. Thus, we trained a dictionary of 32 filters with the fruit dataset beforehand. These filters are of size $16\times 16$, $\Lambda=10$,  and $\tau = 0.06$. It's worth noting that the computation of convolution is finished in the Fourier domain, which can be regarded as dot product, thus the size of filters can be even. Fig .\ref{fig:dictionary} shows the trained dictionary in visual.
\subsection{Color Images}

Before the experiments begin, we execute the missing process to original images. Specifically, to obtain observed data, all the pixels in color images will be erased randomly according to a setting ratio (same in the following MRI and video experiments).

Fig. \ref{fig:colorimage} shows the recovered results of color images by all comparing models. The first six images are  $256\times 256\times 3$ in size, while the last two images are $256\times 192\times 3$ in size. The details are marked out and enlarged in the mark boxes. Comparing to the degradation models of the proposed ones, HaLRTC and LRTC-TV, it's intuitionistic to show the effect of CSC prior. The Framelet-TV method is very competitive to our model LRTC-CSC-I, however, the proposed model holds more clean and smooth details even in an inferior comparison on last two color images. In most situations, LRTC-CSC-I can obtain a higher performance. And the results of LRTC-CSC-II are the most similar to the original images, especially.

The PSNR and SSIM values of recovered results for the 8 color images by different algorithms are shown in Tab.\ref{tab:psnr} and \ref{tab:ssim}. One can see that the LRTC-TNN method is superior to SNN-based HaLRTC on the whole. LRTC-TV is very closed to TNN-3DTV on SSIM values but inferior on PSNR values.  And Framelet-TV method holds onto the position of the third place by a comfortable margin in most situations. Its performance ends up second only to the proposed two models. Especially, the LRTC-CSC models have a larger lead at color images similar to peppers and fruits due to the dictionary trained by a fruit dataset, which also shows the importance of the relationship between underlying data and dataset. Thus why the results of LRTC-CSC-I on last two images are inferior to Framelet-TV can be explained by the few relationships between underlying tensors and dataset. In this way, the CSC plays a small role. Besides, when the missing ratio is 90\%, other algorithms can hardly hold a stable SSIM value (especially the last three images), LRTC-CSC models are still standing at a relatively high level.

\subsection{MRI dataset}

In this subsection, an MRI dataset of size $180\times216\times30$ is chosen to test our model. Except for the 30 bands used for the test, we select another 10 bands from the raw data as the dictionary training set. And different from the color image experiments, the dictionary used in this part is of size $16\times16\times30$ for working well in most situations.

\begin{figure*}[htbp]
	\centering
	\subfigure[]{
		\begin{minipage}[b]{0.095\textwidth}
			\centering
			\footnotesize Original\vspace{2pt}
			\includegraphics[width=\textwidth]{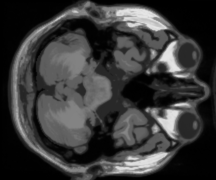}\vspace{2pt}
			\includegraphics[width=\textwidth]{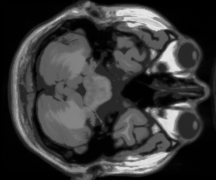}\vspace{2pt}
			\includegraphics[width=\textwidth]{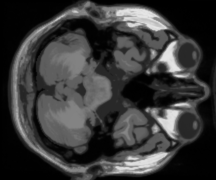}\vspace{2pt}
		\end{minipage}
	}
	\hspace{-2.5ex}
	\subfigure[]{
		\begin{minipage}[b]{0.095\textwidth}
			\centering
			\footnotesize Observed\vspace{2pt}
			\includegraphics[width=\textwidth]{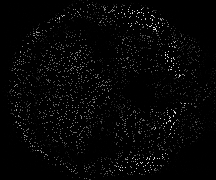}\vspace{2pt}
			\includegraphics[width=\textwidth]{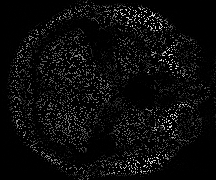}\vspace{2pt}
			\includegraphics[width=\textwidth]{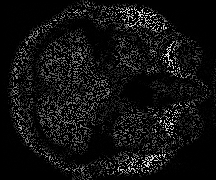}\vspace{2pt}
		\end{minipage}
	}
	\hspace{-2.5ex}
	\subfigure[]{
		\begin{minipage}[b]{0.095\textwidth}
			\centering
			\footnotesize SNN\vspace{2pt}
			\includegraphics[width=\textwidth]{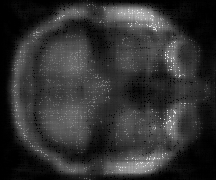}\vspace{2pt}
			\includegraphics[width=\textwidth]{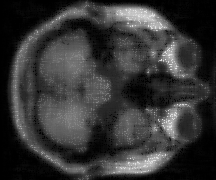}\vspace{2pt}
			\includegraphics[width=\textwidth]{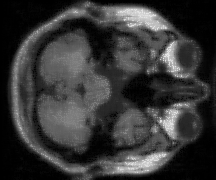}\vspace{2pt}
		\end{minipage}
	}
	\hspace{-2.5ex}
	\subfigure[]{
		\begin{minipage}[b]{0.095\textwidth}
			\centering
			\footnotesize TNN\vspace{2pt}
			\includegraphics[width=\textwidth]{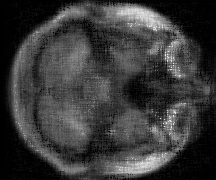}\vspace{2pt}
			\includegraphics[width=\textwidth]{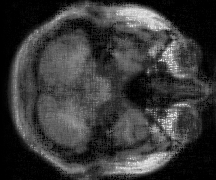}\vspace{2pt}
			\includegraphics[width=\textwidth]{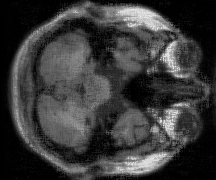}\vspace{2pt}
		\end{minipage}
	}
	\hspace{-2.5ex}
	\subfigure[]{
		\begin{minipage}[b]{0.095\textwidth}
			\centering
			\footnotesize TV\vspace{2pt}
			\includegraphics[width=\textwidth]{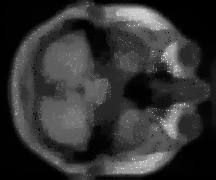}\vspace{2pt}
			\includegraphics[width=\textwidth]{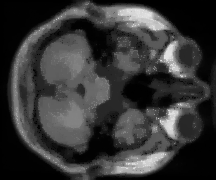}\vspace{2pt}
			\includegraphics[width=\textwidth]{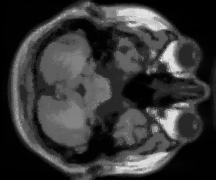}\vspace{2pt}
		\end{minipage}
	}
	\hspace{-2.5ex}
	\subfigure[]{
		\begin{minipage}[b]{0.095\textwidth}
			\centering
			\footnotesize Framelet\vspace{2pt}
			\includegraphics[width=\textwidth]{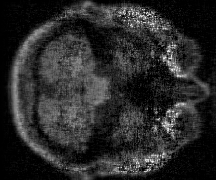}\vspace{2pt}
			\includegraphics[width=\textwidth]{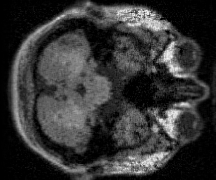}\vspace{2pt}
			\includegraphics[width=\textwidth]{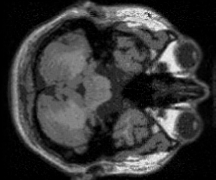}\vspace{2pt}
		\end{minipage}
	}
	\hspace{-2.5ex}
	\subfigure[]{
		\begin{minipage}[b]{0.095\textwidth}
			\centering
			\footnotesize 3DTV\vspace{2pt}
			\includegraphics[width=\textwidth]{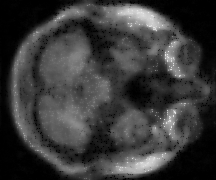}\vspace{2pt}
			\includegraphics[width=\textwidth]{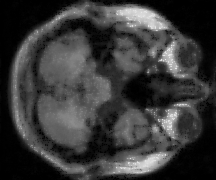}\vspace{2pt}
			\includegraphics[width=\textwidth]{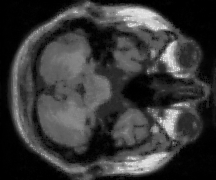}\vspace{2pt}
		\end{minipage}
	}
	\hspace{-2.5ex}
	\subfigure[]{
		\begin{minipage}[b]{0.095\textwidth}
			\centering
			\footnotesize Framelet-TV\vspace{2pt}
			\includegraphics[width=\textwidth]{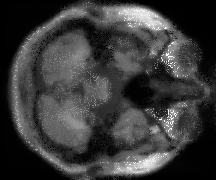}\vspace{2pt}
			\includegraphics[width=\textwidth]{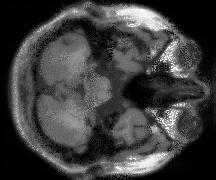}\vspace{2pt}
			\includegraphics[width=\textwidth]{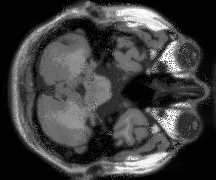}\vspace{2pt}
		\end{minipage}
	}
	\hspace{-2.5ex}
	\subfigure[]{
		\begin{minipage}[b]{0.095\textwidth}
			\centering
			\footnotesize CSC-I\vspace{2pt}
			\includegraphics[width=\textwidth]{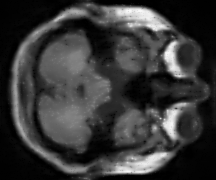}\vspace{2pt}
			\includegraphics[width=\textwidth]{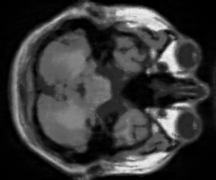}\vspace{2pt}
			\includegraphics[width=\textwidth]{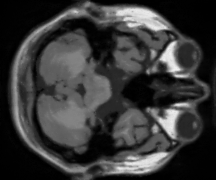}\vspace{2pt}
		\end{minipage}
	}
	\hspace{-2.5ex}
	\subfigure[]{
		\begin{minipage}[b]{0.095\textwidth}
			\centering
			\footnotesize CSC-II\vspace{2pt}
			\includegraphics[width=\textwidth]{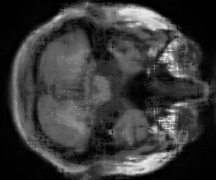}\vspace{2pt}
			\includegraphics[width=\textwidth]{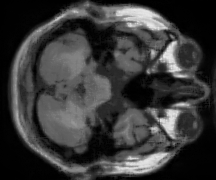}\vspace{2pt}
			\includegraphics[width=\textwidth]{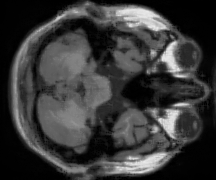}\vspace{2pt}
		\end{minipage}
	}
	\hspace{-2.5ex}
	\caption{The 30-th band of recovered MRI data by all algorithms for different sampling rates. From (a) to (j): Original data, Observed data, recovered data by SNN, TNN, TV, Framelet, 3DTV, Framelet-TV, CSC-I and CSC-II, respectively. From the top down:  10\%, 20\% and 30\% sampling rate.}
	\label{fig:mri}
\end{figure*}
\begin{figure*}[htbp]
	\centering
	\subfigure[]{
		\begin{minipage}[b]{0.095\textwidth}
			\centering
			\footnotesize Original\vspace{2pt}
			\includegraphics[width=\textwidth]{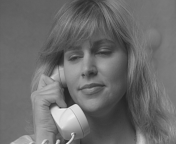}\vspace{2pt}
			\includegraphics[width=\textwidth]{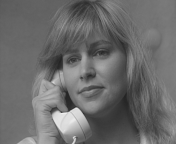}\vspace{2pt}
			\includegraphics[width=\textwidth]{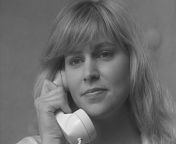}\vspace{2pt}
		\end{minipage}
	}
	\hspace{-2.5ex}
	\subfigure[]{
		\begin{minipage}[b]{0.095\textwidth}
			\centering
			\footnotesize Observed\vspace{2pt}
			\includegraphics[width=\textwidth]{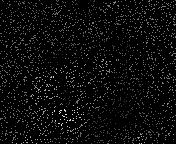}\vspace{2pt}
			\includegraphics[width=\textwidth]{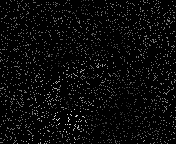}\vspace{2pt}
			\includegraphics[width=\textwidth]{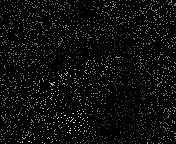}\vspace{2pt}
		\end{minipage}
	}
	\hspace{-2.5ex}
	\subfigure[]{
		\begin{minipage}[b]{0.095\textwidth}
			\centering
			\footnotesize SNN\vspace{2pt}
			\includegraphics[width=\textwidth]{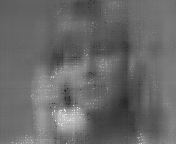}\vspace{2pt}
			\includegraphics[width=\textwidth]{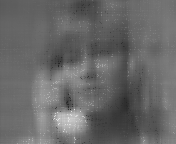}\vspace{2pt}
			\includegraphics[width=\textwidth]{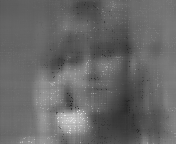}\vspace{2pt}
		\end{minipage}
	}
	\hspace{-2.5ex}
	\subfigure[]{
		\begin{minipage}[b]{0.095\textwidth}
			\centering
			\footnotesize TNN\vspace{2pt}
			\includegraphics[width=\textwidth]{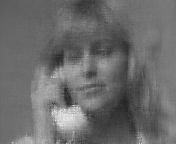}\vspace{2pt}
			\includegraphics[width=\textwidth]{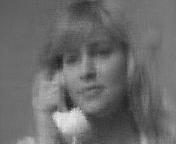}\vspace{2pt}
			\includegraphics[width=\textwidth]{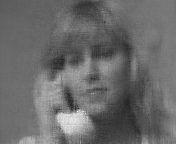}\vspace{2pt}
		\end{minipage}
	}
	\hspace{-2.5ex}
	\subfigure[]{
		\begin{minipage}[b]{0.095\textwidth}
			\centering
			\footnotesize TV\vspace{2pt}
			\includegraphics[width=\textwidth]{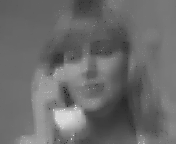}\vspace{2pt}
			\includegraphics[width=\textwidth]{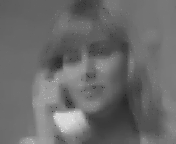}\vspace{2pt}
			\includegraphics[width=\textwidth]{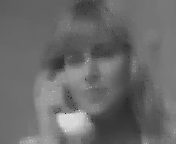}\vspace{2pt}
		\end{minipage}
	}
	\hspace{-2.5ex}
	\subfigure[]{
		\begin{minipage}[b]{0.095\textwidth}
			\centering
			\footnotesize Framelet\vspace{2pt}
			\includegraphics[width=\textwidth]{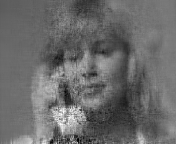}\vspace{2pt}
			\includegraphics[width=\textwidth]{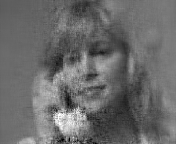}\vspace{2pt}
			\includegraphics[width=\textwidth]{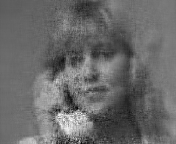}\vspace{2pt}
		\end{minipage}
	}
	\hspace{-2.5ex}
	\subfigure[]{
		\begin{minipage}[b]{0.095\textwidth}
			\centering
			\footnotesize 3DTV\vspace{2pt}
			\includegraphics[width=\textwidth]{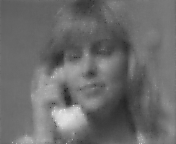}\vspace{2pt}
			\includegraphics[width=\textwidth]{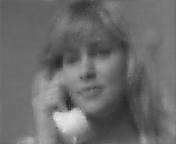}\vspace{2pt}
			\includegraphics[width=\textwidth]{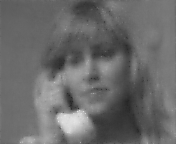}\vspace{2pt}
		\end{minipage}
	}
	\hspace{-2.5ex}
	\subfigure[]{
		\begin{minipage}[b]{0.095\textwidth}
			\centering
			\footnotesize Framelet-TV\vspace{2pt}
			\includegraphics[width=\textwidth]{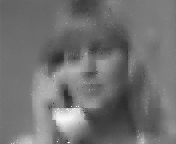}\vspace{2pt}
			\includegraphics[width=\textwidth]{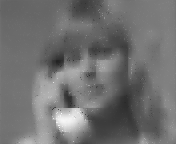}\vspace{2pt}
			\includegraphics[width=\textwidth]{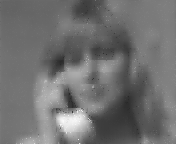}\vspace{2pt}
		\end{minipage}
	}
	\hspace{-2.5ex}
	\subfigure[]{
		\begin{minipage}[b]{0.095\textwidth}
			\centering
			\footnotesize CSC-I\vspace{2pt}
			\includegraphics[width=\textwidth]{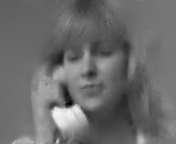}\vspace{2pt}
			\includegraphics[width=\textwidth]{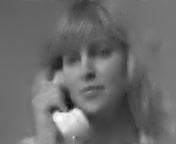}\vspace{2pt}
			\includegraphics[width=\textwidth]{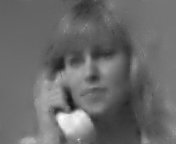}\vspace{2pt}
		\end{minipage}
	}
	\hspace{-2.5ex}
	\subfigure[]{
		\begin{minipage}[b]{0.095\textwidth}
			\centering
			\footnotesize CSC-II\vspace{2pt}
			\includegraphics[width=\textwidth]{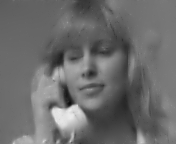}\vspace{2pt}
			\includegraphics[width=\textwidth]{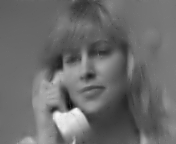}\vspace{2pt}
			\includegraphics[width=\textwidth]{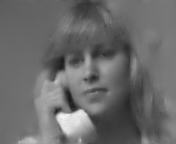}\vspace{2pt}
		\end{minipage}
	}
	\hspace{-2.5ex}
	\caption{The 5-th, 15-th and 25-th frame of recovered \textit{suzie} video by all algorithms for 10\% sampling rates. From (a) to (j): Original data, Observed data, recovered data by SNN, TNN, TV, Framelet, 3DTV, Framelet-TV, CSC-I and CSC-II, respectively. From the top down:  5-th, 15-th and 25-th frame.}
	\label{fig:video}
\end{figure*}

\begin{figure}[tp]
	\centering
	\includegraphics[width=0.23\textwidth]{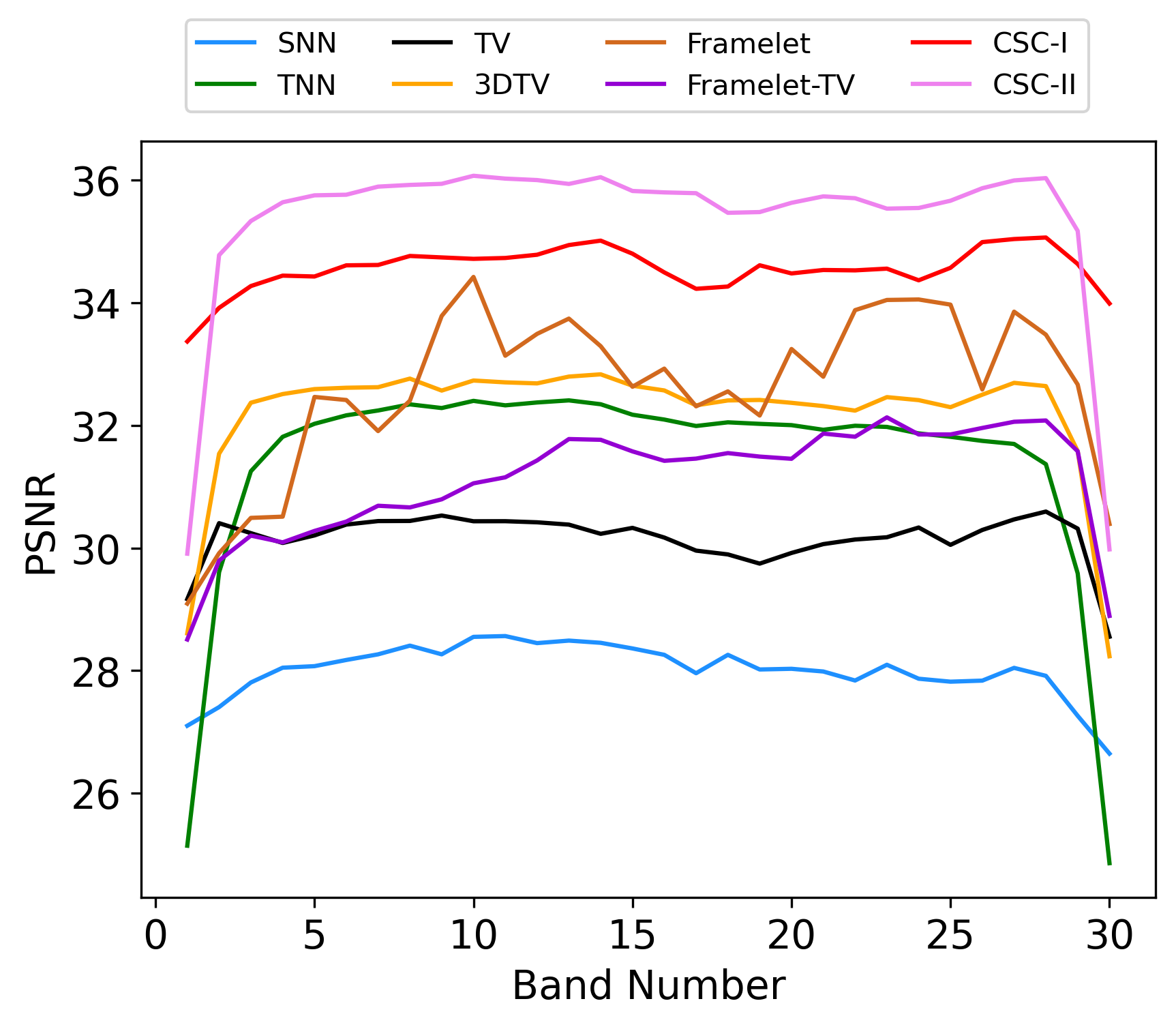}\hspace{-1pt}
	\includegraphics[width=0.235\textwidth]{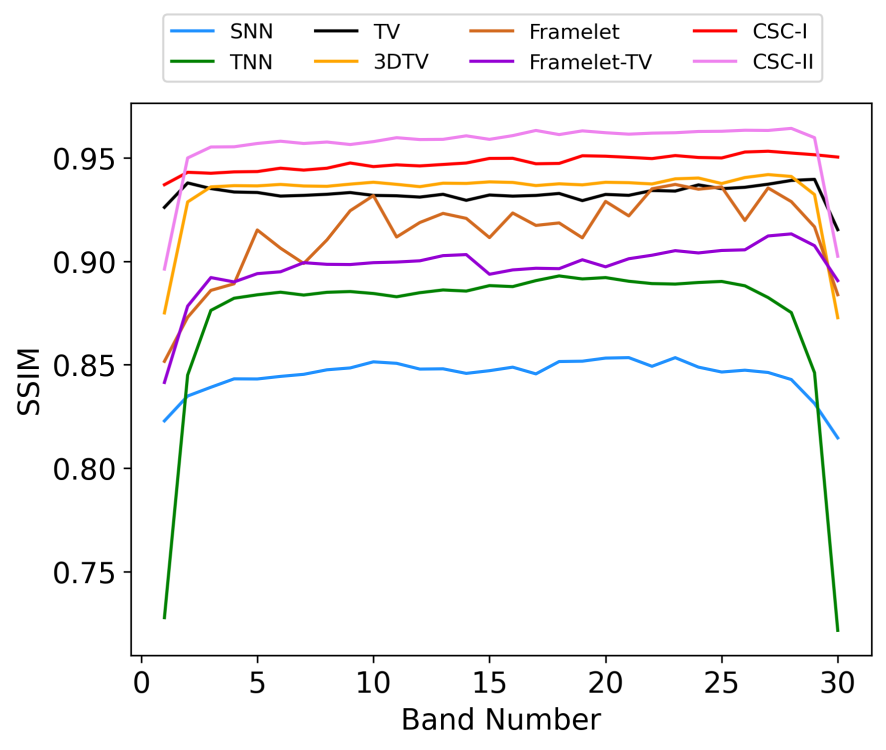}
	\caption{The PSNR and SSIM values of different bands of MRI data recovered by all algorithms for the sampling rate 30\%.}
	\label{fig:mri_band}
\end{figure}


Color images own three channels only, while MRI data often include dozen or hundred bands. In this way, the relationship among different bands will be closer, which is very advantageous to the TNN-based method. However, our SNN-based model still shows its superiority after introducing the CSC prior. The same as color images, we randomly pick 10\%, 20\%, and 30\% samples as the observations, and the results are shown in Fig. \ref{fig:mri}.
It's intuitive that LRTC-CSC models achieve the best performance among these models, while HaLRTC achieves the worst performance, which verified the effectiveness of CSC regularization. In particular, LRTC-CSC models work well on detailed recovery comparing to TNN-3DTV and Framelet-TV. When the missing rate reaches 90\%, our model can still hold a relatively clear result. 

\begin{figure}[tp]
	\centering
	\includegraphics[width=0.23\textwidth]{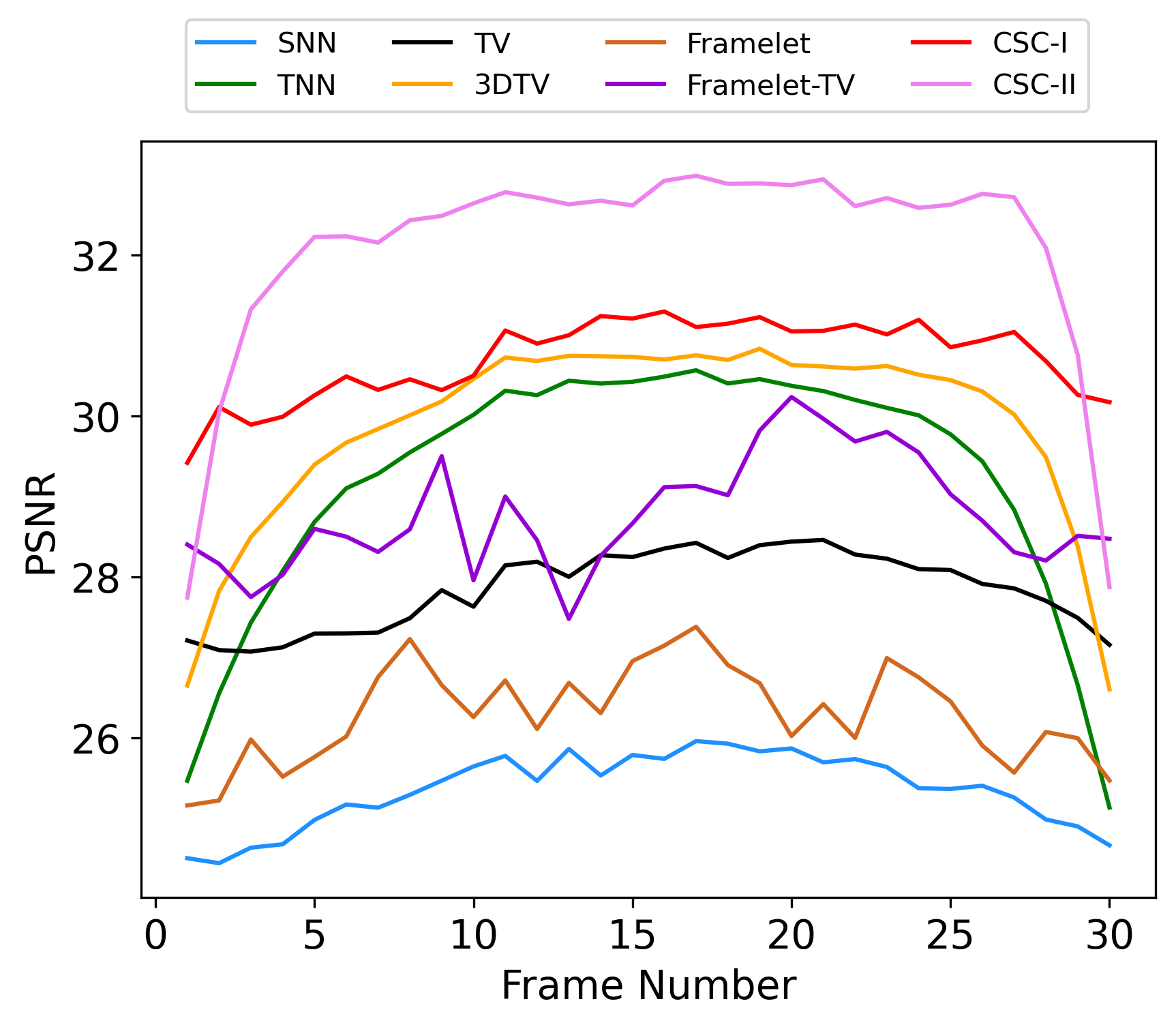}\hspace{-1pt}
	\includegraphics[width=0.235\textwidth]{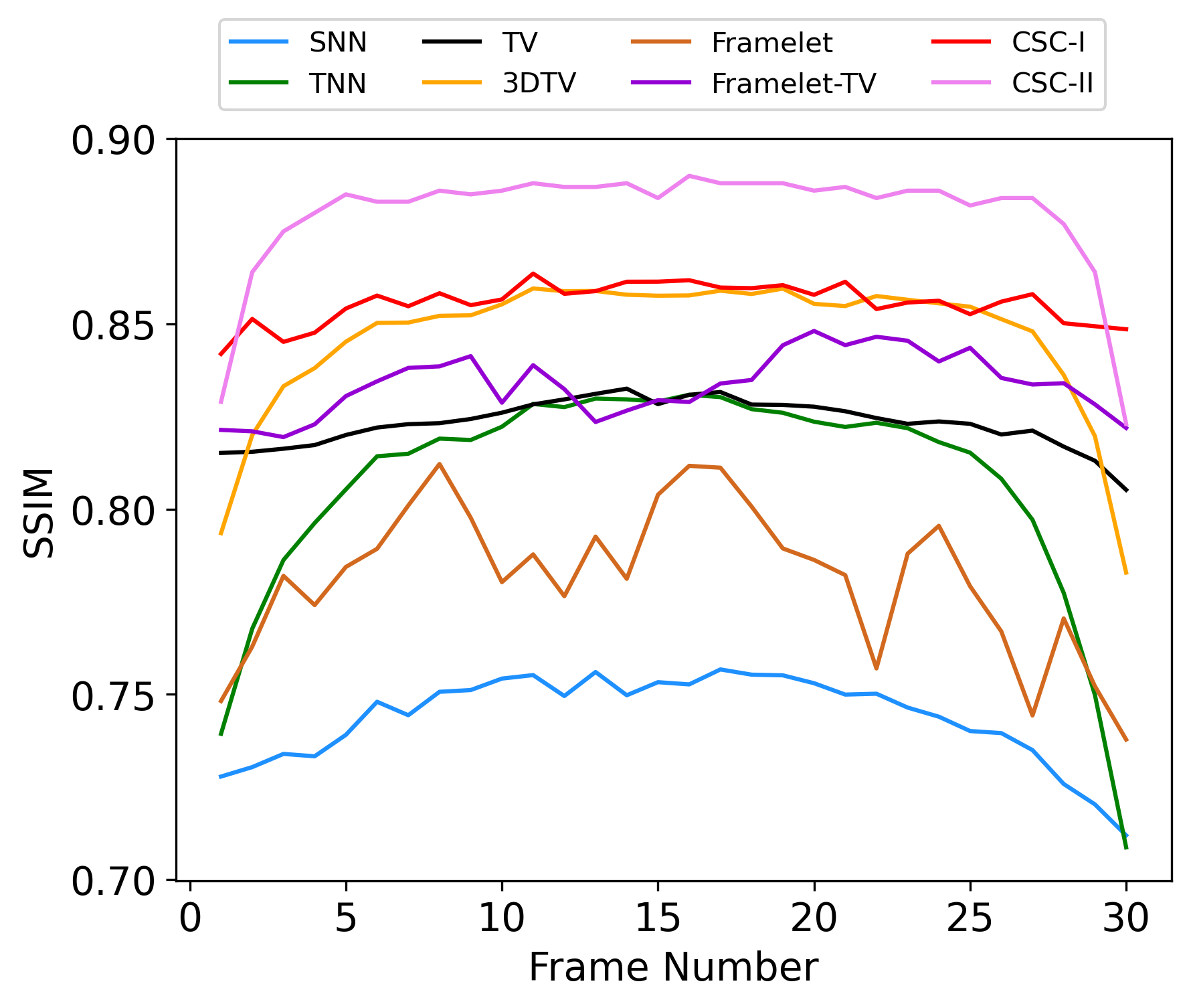}
	\caption{The PSNR and SSIM values of different frames of video data recovered by all algorithms for the sampling rate 10\%.}
	\label{fig:suzie_frame}
\end{figure}

\begin{figure*}[tp]
	\centering
	\subfigure[]{
		\includegraphics[width=0.24\textwidth]{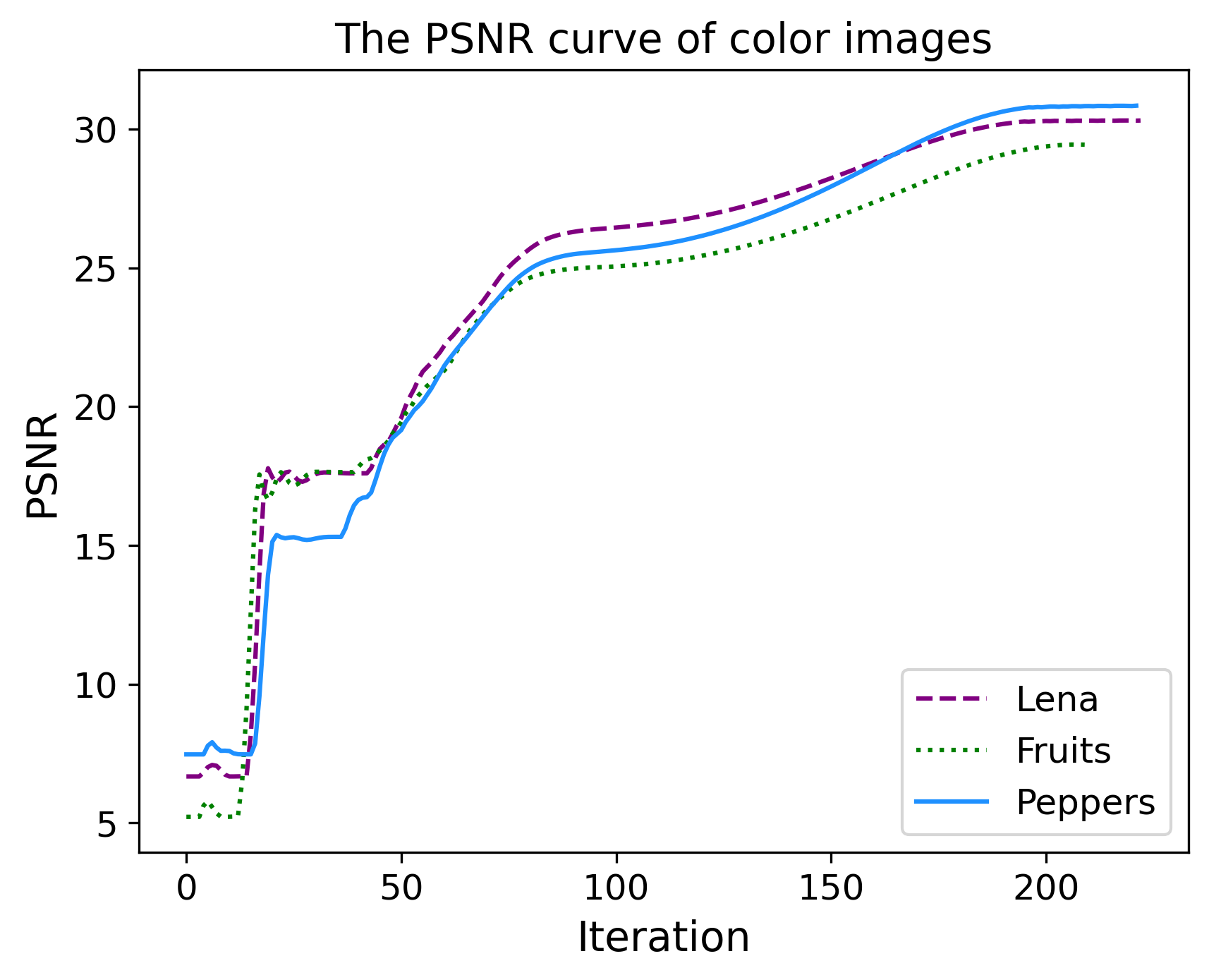}\hspace{-1pt}
		\includegraphics[width=0.24\textwidth]{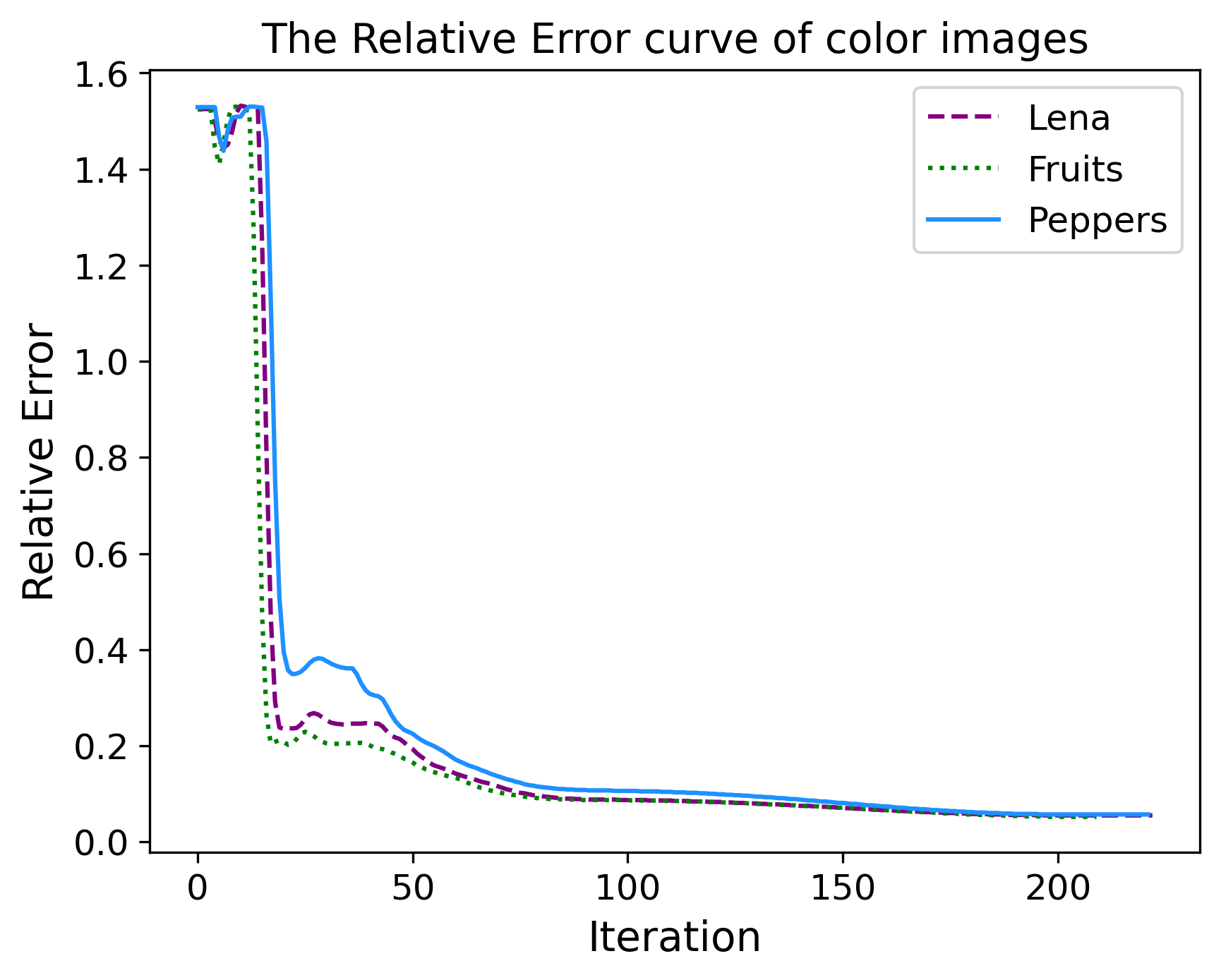}\hspace{-1pt}
		}
	\subfigure[]{
		\includegraphics[width=0.24\textwidth]{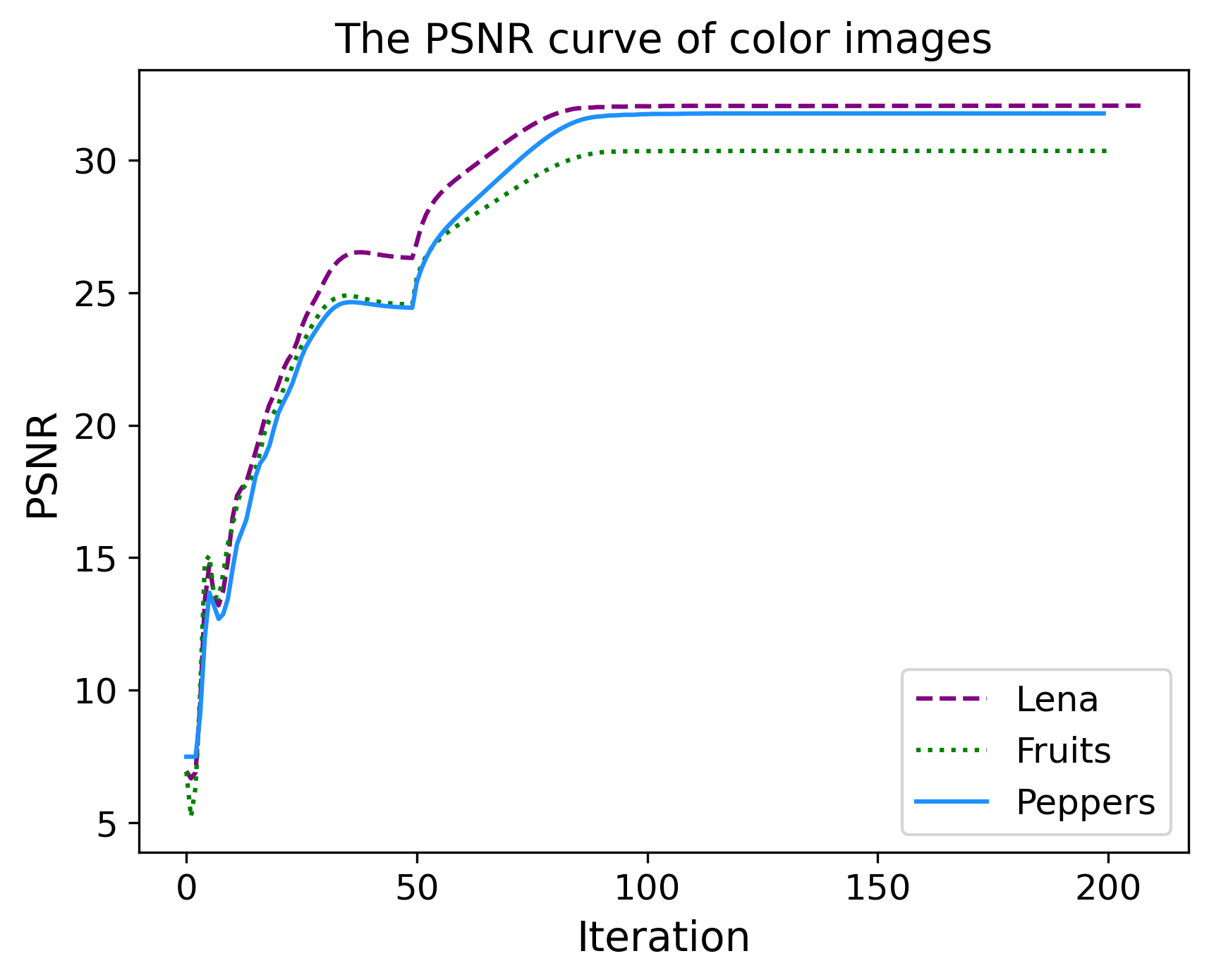}\hspace{-1pt}
		\includegraphics[width=0.24\textwidth]{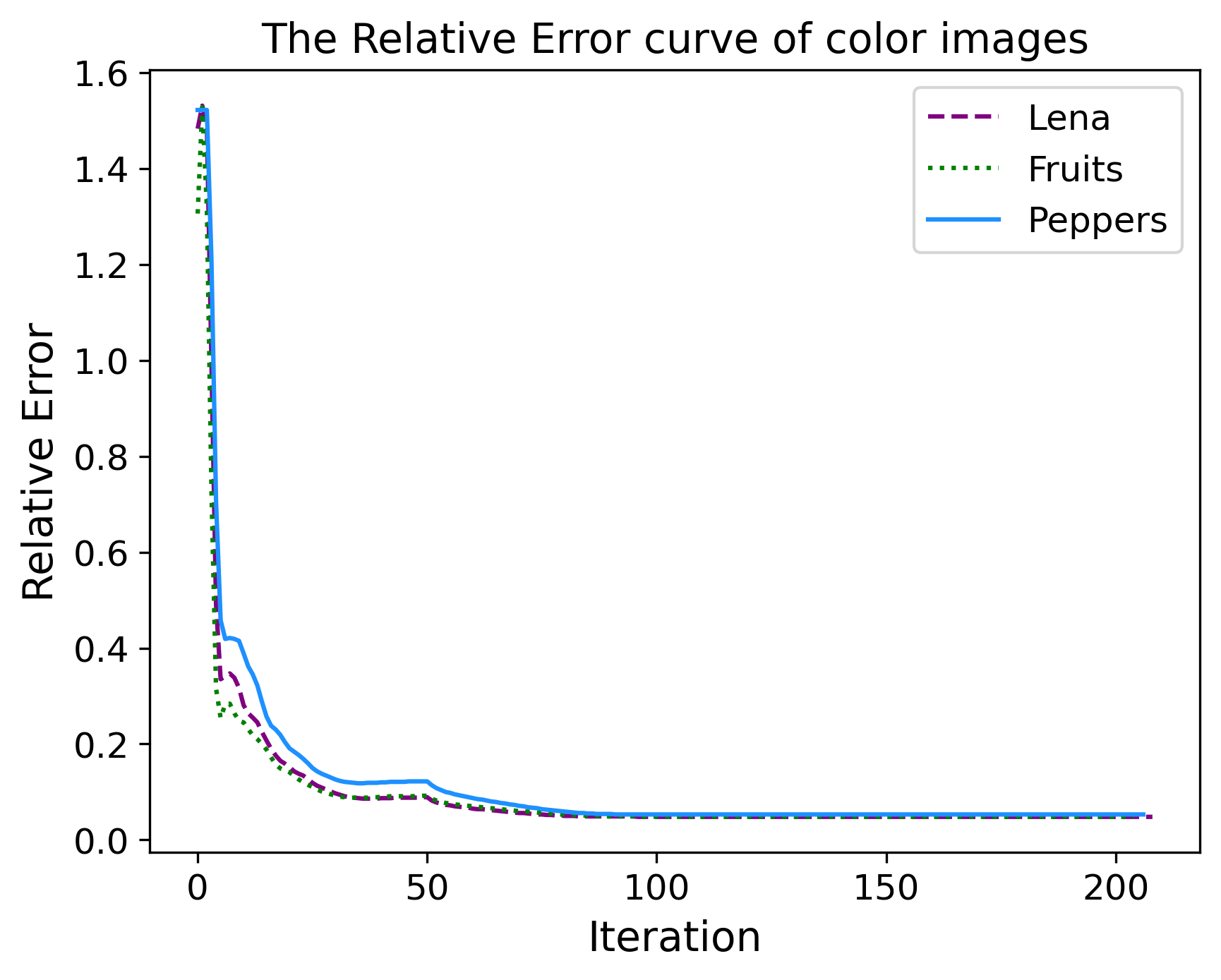}\hspace{-1pt}
	}

	\caption{The convergence behavior with respect to iterations on the color images at $MR=70\%$. The figures show the convergence of CSC-I (a) and CSC-II (b) algorithms for RE and PSNR, respectively. }
	\label{fig:Iteration}
\end{figure*}

Fig. \ref{fig:mri_band} further shows the recovery performance in every band. Both the PSNR and SSIM values of each band show the superiority of LRTC-CSC models. Note that TNN-3DTV can achieve a high SSIM value approaching LRTC-CSC. However, there is a gap between them on the edge band of the whole MRI data. The proposed model LRTC-CSC-I achieves a performance promotion of $4$ and $0.05$ with respect to PSNR and SSIM over the results of TNN-3DTV.
Moreover, the performances of LRTC-TNN and LRTC-CSC-II on edge band are similar to TNN-3DTV, which implies the shortcoming of TNN-based methods. In other words, TNN can not handle the edge information due to lacking enough neighbors, which is determined by its predominant characteristic of global information capturing. The peformance of TT rank-based method is not satisfactory in MRI data recovering experiments because of imposing KA on the unbalanced tensor \cite{yang2020tensor}\cite{chen2021auto}.

\subsection{Videos}

In this subsection, we benchmarked our model on \textit{Suzie} of size $144\times176\times150$. The same as MRI experiments, we utilize 30 filters in a dictionary trained by only 10 frames of the raw video data. And except for the trained ones, another 30 frames of video are used for recovering task. By randomly selecting 10\% samples, the recovered results are shown in Fig. \ref{fig:video}. It is not hard to see that the results recovered by our LRTC-CSC models are the cleanest. Similar to MRI data experiment, TNN-3DTV is close to LRTC-CSC-I but it can not handle the edge frame well, e.g., the 30-th frame. It can also be observed that the SNN-based Framelet-TV is inferior to TNN-3DTV, which shows the low-rank approximating ability of TNN.

Fig. \ref{fig:suzie_frame} shows the recovery results of each frame in detail. Note that LRTC-CSC-II achieves the highest score in both PSNR and SSIM. However, at the beginning or end of the video, only the proposed LRTC-CSC-I is stable while the index curve of the TNN-based methods, including LRTC-CSC-II, drops rapidly, which shows the stability of SNN-based methods on edges, again. Compared with HaLRTC and LRTC-TNN, the proposed models have claimed the superiority of CSC regularization.

\subsection{Convergency analysis}
The numerical experiments have shown the remarkable performance of CSC-I and CSC-II. 
In order to demonstrate the convergence of algorithms in numerical, Fig. \ref{fig:Iteration} shows the variation curves of the Relative Error (RE) values, i.e., $RE=\frac{\Vert \mathcal{X-T}\Vert_F}{\Vert \mathcal{T}\Vert_F}$, and PSNR values of CSC-I and CSC-II algorithms with the number of iterations on three color images of \textit{Lena}, \textit{Peppers} and \textit{Fruits} at $MR=70\%$.
Factly, it can be observed that both the proposed algorithms converge to an optimal point.

\section{Conclusion}
Inspired by the capacity of CSC on high-frequency component processing, we introduced CSC into the traditional LRTC model. In this way, the details of the underlying tensor would be improved in addition to the low-rank component recovery. Comparing to other plug-and-play CNN prior to using thousands of samples, our method can achieve good performance with only a few samples. For the proposed models, we obtain effective algorithms based on the inexact ADMM method. And the effectiveness of LRTC-CSC-I and LRTC-CSC-II have been verified in color images, MRI data, and video data recovery experiments.

In future work, it would be of great interest to leverage the prior information
of CSC to improve the performance of other models. For the marginal effects of the TNN-based method, CSC might be a promising method to strengthen the connection between the edge and subject.

\section*{Acknowledgment}

The work described in this paper was supported by the Key-Area Research and Development Program of Guangdong Province (2018B010109001), the National Natural Science Foundation of China (11801595), the Guangdong Basic and Applied Basic Research Foundation (2019A1515011043), the Natural Science Foundation of Guangdong (2018A030310076) and Tencent Wechat Rhino-bird project No.2021321.



%
\bibliographystyle{IEEEtran}
\bibliography{references}

\begin{IEEEbiography}[{\includegraphics[width=1in,height=1.1in]{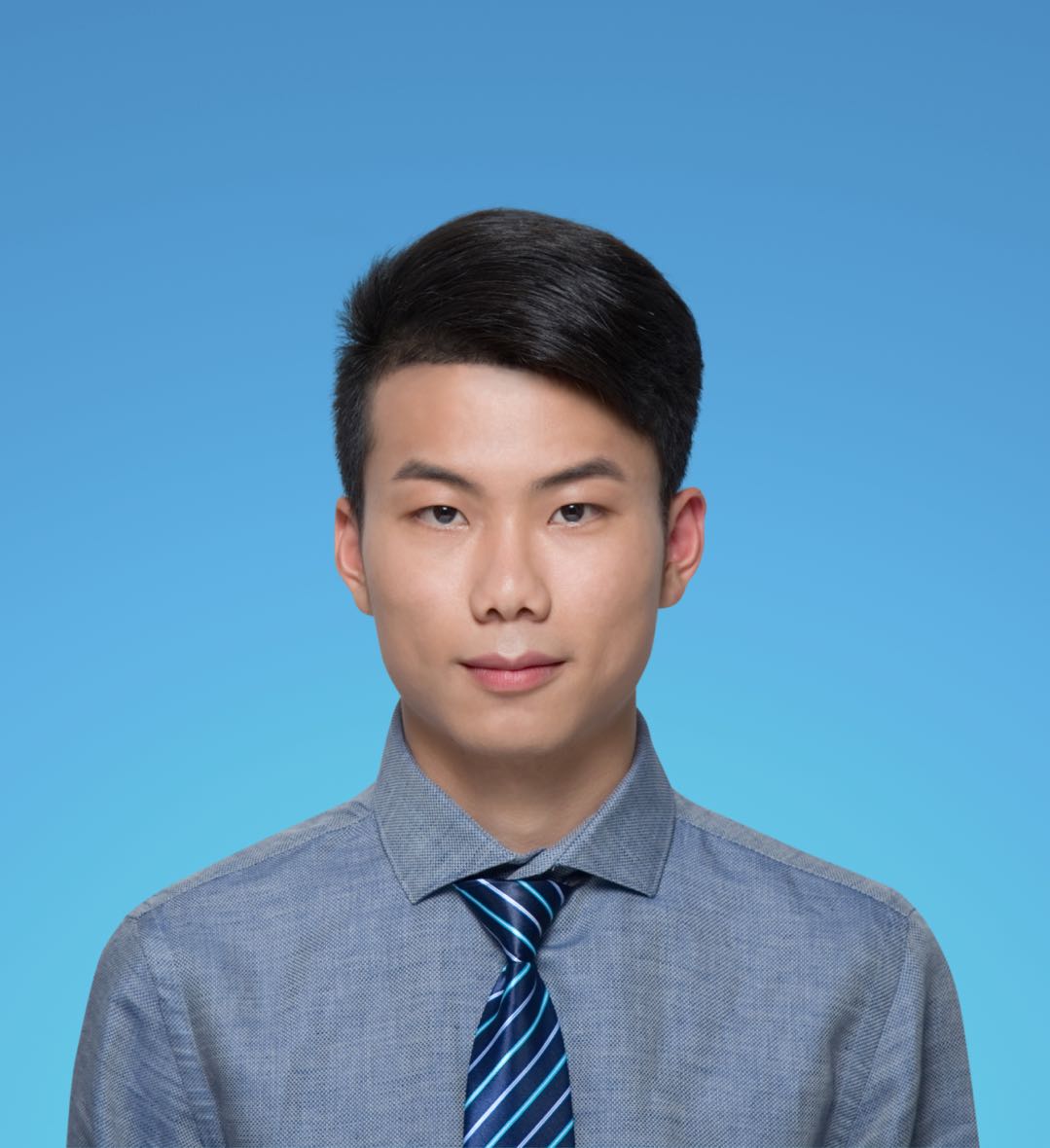}}]{Zhebin Wu}
	received the B.S. degree majoring in Information and Computing Science in Sun Yat-sen University, Guangzhou, China, in 2020. And now he is pursuing the Master degree in Sun Yat-sen University. His current research interests focus on numerical optimization, numerical linear algebra and machine learning.
\end{IEEEbiography}

\begin{IEEEbiography}[{\includegraphics[width=1in,height=1.3in]{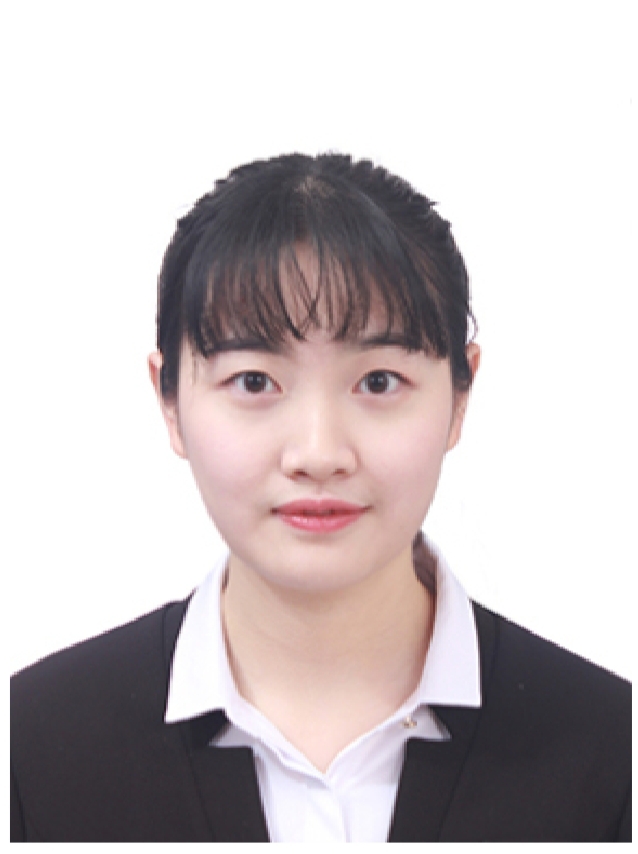}}]{Tianchi Liao}
	received the B.S. degree majoring in Information and Computing Science in Sichuan Agricultural University, Sichuan, China, in 2021. And now she is pursuing the Master degree in Sun Yat-sen University. Her research interests include models and algorithms of low-rank modeling for image processing problems.
\end{IEEEbiography}

\begin{IEEEbiography}[{\includegraphics [width=1in,height=1.25in] {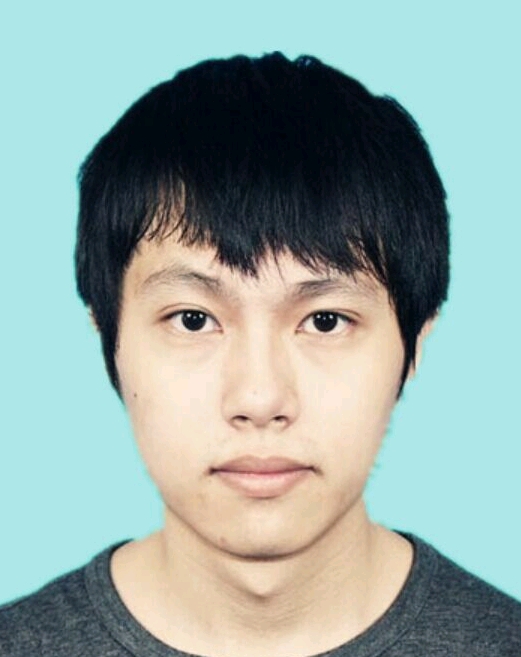}}] {Chuan Chen}
	received the B.S. degree from SunYat-sen University, Guangzhou, China, in 2012, andthe Ph.D. degree from Hong Kong Baptist University, Hong Kong, in 2016.  He is currently a Research Associate Professor with the School of Data and Computer Science, SunYat-Sen University. His current research interestsinclude machine learning, numerical linear algebra,and numerical optimization. 
\end{IEEEbiography}

\begin{IEEEbiography}[{\includegraphics [width=1in,height=1.25in] {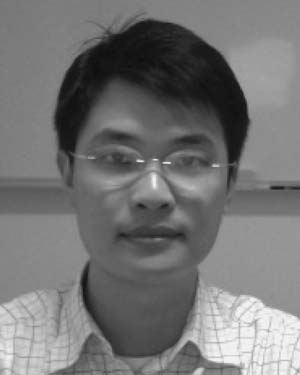}}] {Cong Liu}
	received the B.S. degree in microelectronics from South China University of Technology,
	Guangzhou, China, in 2002; the M.S. degree in computer software and theory from Sun Yat-Sen University, Guangzhou, in 2005; and the Ph.D. degree in computer science from Florida Atlantic University, Boca Raton, FL, USA.
	
	He is currently an Assistant Professor with the Department of Computer Science, Sun Yat-Sen University. His main research interests include delay-tolerant
	networks and deep packet inspection.
\end{IEEEbiography}

\begin{IEEEbiography}[{\includegraphics [width=1in,height=1.1in] {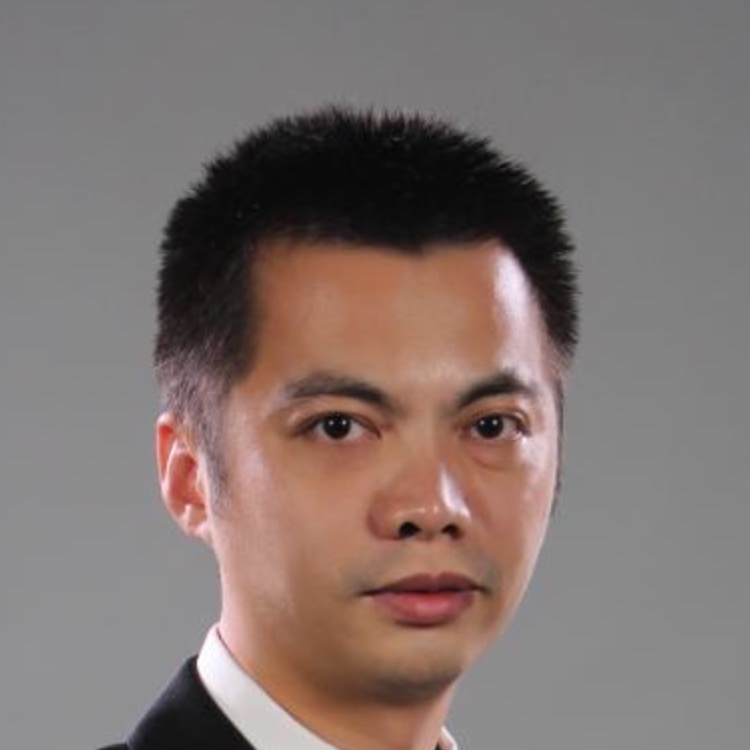}}] {Zibin Zheng}
	received the Ph.D. degree from the Chinese University of Hong Kong, in 2011.
	He is currently a Professor at School of Data and Computer Science with Sun Yat-sen University, China. He serves as Chairman of the Software Engineering Department. He published over 120 international journal and conference papers, including 3 ESI highlycited papers. According to Google Scholar, his papers have more than 7000 citations, with an H-index of 42. His research interests include blockchain, services computing, software engineering, and financial big data. He was a recipient of several awards, including the Top 50 Influential Papers in Blockchain of 2018, the ACM
	SIGSOFT Distinguished Paper Award at ICSE2010, the Best Student Paper Award at ICWS2010. He served as BlockSys'19 and CollaborateCom’16 General Co-Chair, SC2'19, ICIOT’18 and IoV’14 PC Co-Chair.
\end{IEEEbiography}

\begin{IEEEbiography}[{\includegraphics [width=1in,height=1.25in] {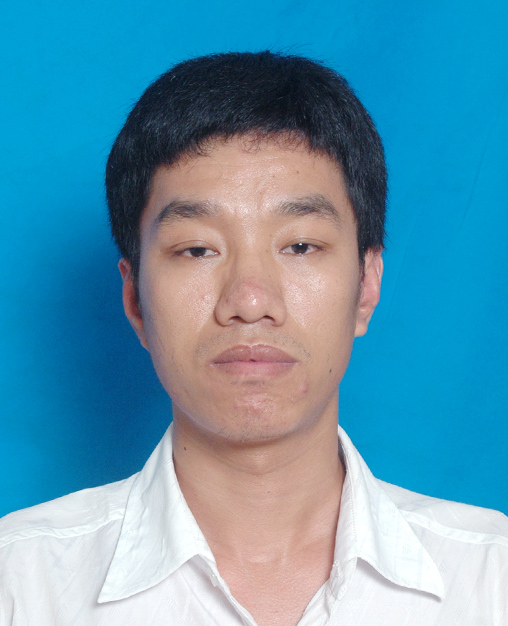}}] {Xiongjun Zhang}
	received the Ph.D. degree from the
	College of Mathematics and Econometrics, Hunan
	University, Changsha, China, in 2017.
	From 2015 to 2016, he was an Exchange
	Ph.D. Student with the Department of Mathematics, Hong Kong Baptist University, Hong Kong.
	He is currently an Assistant Professor with the
	School of Mathematics and Statistics, Central China
	Normal University, Wuhan, China. His current
	research interests include image processing and
	tensor optimization.
\end{IEEEbiography}

\end{document}